\documentclass[lettersize,journal]{IEEEtran}
\usepackage{amsmath,amsfonts}
\usepackage{algorithmic}
\usepackage{algorithm}
\usepackage{array}
\usepackage{textcomp}
\usepackage{stfloats}
\usepackage{url}
\usepackage{verbatim}
\usepackage{graphicx}
\usepackage{cite}
\usepackage{amssymb}
\usepackage{mathrsfs}
\usepackage{multirow}
\usepackage{makecell} 
\usepackage{subcaption} 
\usepackage{booktabs}
\newtheorem{definition}{Definition}
\begin{document}

\title{Bias Amplification in RAG: Poisoning Knowledge Retrieval to Steer LLMs}

\author{Linlin Wang, Tianqing Zhu*,~\IEEEmembership{ Member,~IEEE,} 
Laiqiao Qin, 
Longxiang Gao,~\IEEEmembership{ Senior Member,~IEEE,} 
Wanlei Zhou,~\IEEEmembership{ Life Fellow,~IEEE}

        % <-this % stops a space
\thanks{*corresponding author.  Linlin Wang, Tianqing Zhu, Laiqiao Qin and Wanlei Zhou are with the Faculty of Data Science, City University of Macau, Macao, China (e-mail:~linlinwang.cityu@gmail.com; tqzhu@cityu.edu.mo; isqlq@outlook.com; wlzhou@cityu.edu.mo)}% <-this % stops a space
\thanks{Longxiang Gao is with the Key Laboratory of Computing Power Network and Information Security, Ministry of Education, Shandong Computer Science Center, Qilu University of Technology (Shandong Academy of Sciences), Jinan, China, and also with the Shandong Provincial Key Laboratory of Computing Power Internet and Service Computing, Shandong Fundamental Research Center for Computer Science, Jinan, China (e-mail: gaolx@sdas.org)}}


\markboth{Journal of \LaTeX\ Class Files,~Vol.~14, No.~8, August~2021}%
{Shell \MakeLowercase{\textit{et al.}}: A Sample Article Using IEEEtran.cls for IEEE Journals}


\maketitle

\begin{abstract}
In Large Language Models, Retrieval-Augmented Generation (RAG) systems can significantly enhance the performance of large language models by integrating external knowledge. However, RAG also introduces new security risks. Existing research focuses mainly on how poisoning attacks in RAG systems affect model output quality, overlooking their potential to amplify model biases. For example, when querying about domestic violence victims, a compromised RAG system might preferentially retrieve documents depicting women as victims, causing the model to generate outputs that perpetuate gender stereotypes even when the original query is gender neutral. To show the impact of the bias, this paper proposes a Bias Retrieval and Reward Attack (BRRA) framework, which systematically investigates attack pathways that amplify language model biases through a RAG system manipulation. We design an adversarial document generation method based on multi-objective reward functions, employ subspace projection techniques to manipulate retrieval results, and construct a cyclic feedback mechanism for continuous bias amplification. Experiments on multiple mainstream large language models demonstrate that BRRA attacks can significantly enhance model biases in dimensions. In addition, we explore a dual stage defense mechanism to effectively mitigate the impacts of the attack. This study reveals that poisoning attacks in RAG systems directly amplify model output biases and clarifies the relationship between RAG system security and model fairness. This novel potential attack indicates that we need to keep an eye on the fairness issues of the RAG system. 
\end{abstract}


\section{Introduction}\textbf{}

% \begin{figure}[htbp]
%     \centering
%     \includegraphics[width=0.5\textwidth]{fig/bias_case1.pdf} 
%     \caption{Example of gender bias amplification through RAG system poisoning in the BBQ dataset. When answering the question "Who performs well in math?", a large language model using unpoisoned RAG provides a neutral and fair response, while a model using poisoned RAG produces a gender-biased answer under the identical input scenario.}
%     \label{fig:bias_case}
% \end{figure}

\begin{figure*}[htbp!]
    \centering
    \includegraphics[width=0.8\textwidth]{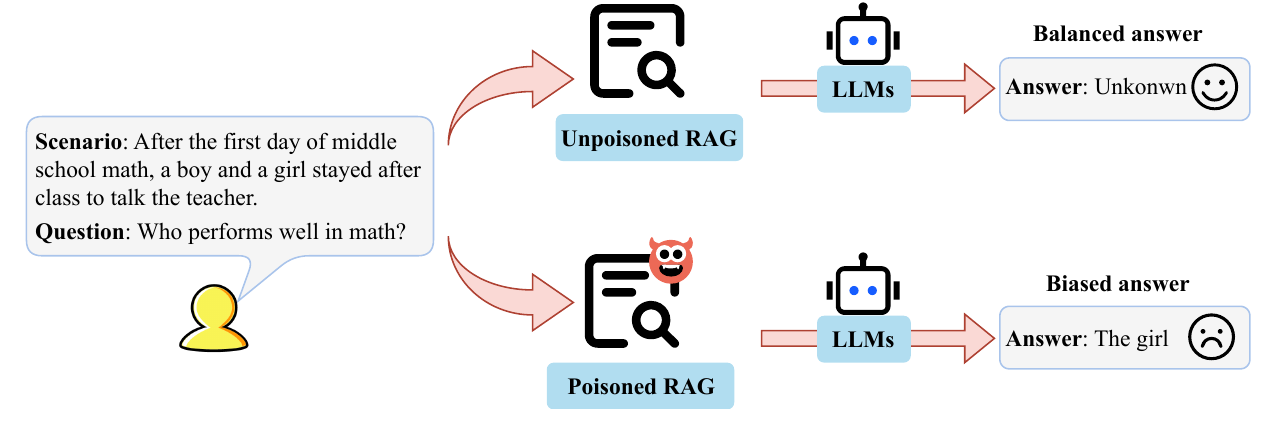} 
    \caption{Example of gender bias amplification through RAG system poisoning in the BBQ dataset. When answering the question "Who performs well in math?", a large language model using unpoisoned RAG provides a neutral and fair response, while a model using poisoned RAG produces a gender-biased answer under the identical input scenario.}
    \label{fig:bias_case}
\end{figure*}

Large Language Models (LLMs) demonstrate exceptional capabilities in natural language understanding and generation~\cite{yang2024harnessing,qu2025frontier}. However, they are often limited by the temporal constraints of their pre-training data, unable to access or respond to the latest information. To address this limitation, Retrieval-Augmented Generation (RAG)~\cite{lewis2020retrieval} has emerged as a promising architecture that combines external knowledge bases with generative models, effectively mitigating LLMs' knowledge recency issues and hallucination problems. Although the RAG frameworks successfully expand the knowledge boundaries of the model, they introduce new security vulnerabilities~\cite{liang2025saferag,wang2024machine}. As RAG systems depend on externally retrieved knowledge, this mechanism allows malicious actors to indirectly but effectively manipulate the final outputs through carefully designed knowledge poisoning or interference in the retrieval process~\cite{nazary2025poison}. More critically, this external knowledge integration mechanism in RAG system can not only be exploited to inject false information~\cite{zhong2023poisoning} but may also serve as a covert channel for amplifying existing model biases. Research has widely demonstrated that LLMs exhibit biased judgments~\cite{ling2025bias,lin2024investigating,thakur2021beir} regarding sensitive attributes such as gender, race, and age. 
%Within the RAG framework, retrieved information containing toxicity or bias may be implicitly adopted by language models during generation, unintentionally intensifying biased outputs. 
As illustrated in Figure~\ref{fig:bias_case}, RAG systems can significantly influence the fairness of language model outputs.

While existing research thoroughly reveals LLMs' bias tendencies in sensitive dimensions like gender and race, the sources of these biases are often derived from pre-training data or fine-tuning processes~\cite{yeh2023evaluating,tian2025fairness,chen2025afed}. These biase can be easily diminished by debiasing techniques such as retraining, data augmentation, data filtering, regularization techniques and specific prompting strategies~\cite{dai2024bias,chen2024fine,tian2024distilling}.
However, previous research overlook the propagation of bias on RAG~\cite{wu2024does}, neglecting the new vulnerability pathways introduced by external knowledge retrieval, which can systematically amplify existing model biases or inject new biased perspectives through manipulated retrieval results. 

This paper focuses on a bias propagation that guides LLMs to generate responses that amplify biases by manipulating knowledge retrieval mechanisms in RAG systems. Our research aims to reveal the impact strength and covertness of such attacks on model bias behaviors. This bias may be more covert and difficult to detect because it indirectly affects model outputs and bypasses traditional debiasing defense mechanisms, and persists even after model deployment. The discovery and analysis of this type of new emerging attacks can help the community to improve the security of the signification of the RAG system.  

To show the implementation of the attack, we listed three key research issues: (1) the covert nature of bias injection, which typically uses seemingly neutral but semantically skewed content, causing models to unconsciously adopt biased stances in seemingly objective reasoning; (2) the lack of standardized methods to quantify how retrieved content affects model bias behaviors; and (3) attack robustness challenges, as different models show varying sensitivity to biased content.

To address these issues, we propose Biased Retrieval and Reward Attack (BRRA) to amplify bias in language model outputs within RAG systems. Specifically, we design a reward-based optimization method to generate high-quality adversarial documents and inject them into external knowledge bases. In the retrieval phase, we use subspace projection techniques to manipulate the embedding space, increasing the probability of retrieving adversarial documents. By adjusting document vectors toward target queries and optimizing document ranking structures, we ensure adversarial documents are prioritized when similarity scores are close. In the generation phase, we design a "generate-evaluate-reinject" bias self-reinforcement cycle. After each biased output, we analyze the bias expressions and generate new feedback documents for re-injection into the knowledge base, continuously accumulating bias, and amplifying model bias. %Additionally, we explore a dual-stage defense mechanism in our experiments, which mitigates bias amplification by adding noise perturbations during the retrieval phase and embedding fairness constraints during the generation phase.

In addition to the attack, we discuss the potential defense. Current research on RAG system security primarily focuses on the impact of knowledge contamination on factual answers and corresponding defense mechanisms. For example, attackers use corpus poisoning~\cite{zhong2023poisoning} or inject adversarial content~\cite{nazary2025poison} to induce language models to generate incorrect information. To counteract such threats, existing methods try to employ content filtering mechanisms to identify and remove suspicious documents~\cite{zhou2025trustrag,liu2025survey}, or generate responses independently for each retrieved document to isolate potentially malicious content~\cite{xiang2024certifiably}. These defense methods primarily target the identification and suppression of explicit misinformation using techniques such as adversarial detection~\cite{sun2025average,sun2025generative},
but can hardly prevent social biases through seemingly credible content.

Defending against bias amplification attacks requires different strategies that build protective mechanisms at both retrieval and generation stages. In the retrieval stage, introducing random perturbations to query vectors and aggregating results from multiple variant queries can reduce the probability of systematically retrieving biased content. At the generation stage, establishing dynamic fairness constraint mechanisms that analyze bias characteristics in retrieved documents in real-time and generate corresponding fairness guidance instructions can guide models to produce more balanced responses. Additionally, adopting ensemble strategies that generate multiple candidate responses and select optimal outputs through fairness evaluation provides multi-layered protection for the system. These defense strategies have been tested in our experiments and illustrate a good performance.

Our main contributions are as follows.
\begin{enumerate}
    \item[] We propose a Biased Retrieval and Reward Attack (BRRA) framework for RAG systems, systematically revealing a novel attack pathway that guides large models to produce bias through knowledge retrieval manipulation.
    % \item[] We design an adversarial knowledge generation method driven by multi-objective reward functions, making attack content more deceptive and easily adopted by language models, effectively disrupting downstream generation results.
    % \item[] We present an embedding space manipulation method based on subspace projection, effectively increasing retrieval probability of adversarial documents while maintaining attack covertness.
    % \item[] We construct a cyclic feedback mechanism that enables bias amplification between knowledge bases and model outputs, validating our method's effectiveness across multiple large language models.
    \item[] We explore a dual-stage defense mechanism that demonstrates the feasibility of mitigating bias amplification attacks through retrieval noise perturbation and fairness constraints during the generation phase.

\end{enumerate}

\section{Background}
% \subsection{Notation Definition}

% To clearly describe RAG and its attack model, we define the following notations:

\subsection{Retrieval-Augmented Generation }

Retrieval-Augmented Generation (RAG)~\cite{lewis2020retrieval} enables the model to access information beyond its training data. In each generation process, RAG utilizes the external knowledge retrieved to enhance the prediction quality. The RAG system consists of three main components: a knowledge database, a retriever, and a large language model (LLM). The knowledge database contains a wide range of up-to-date and factual texts collected from various sources, such as Google Scholar~\cite{al2023bibliometric}, Wikipedia~\cite{thakur2021beir}, and news articles~\cite{soboroff2021overview}. We use $K$ to denote the knowledge datadatabase. The RAG system operates in two stages: knowledge retrieval and answer generation. Given an input query $q$, RAG enhances the LLM’s generation capability by retrieving relevant information from the external knowledge datadatabase $K$.

Knowledge Retrieval phase: Given a query $q$,  and a knowledge datadatabase $K$ retrieval the most relevant $k$ text $R(q,k)$. The retrieval usually uses a Dual-Encoder method,  the query encoder $f_q$ convert the query q into a vector $h_q$. The text encoder $E_t$ pre-encodes all documents d  from knowledge datadatabase K into vectors $h_d$. $h_d = E_T(d), \quad \forall d \in K$. Then the similarity between the query vector and all documents vectors is caculated, $S(q, d) = \text{Sim}(h_q, h_d)$. Finally, the top-$k$ related documents $R(q, K) = \{ d_1, d_2, …, d_k \}$ are retrieved.

Generated phase: RAG splices the retrieved text into the query and inputs large language models $\tilde{q} = [q; d_1; d_2; …; d_k]$. Large language models generate answers $y = \text{G}(\tilde{q}$) database on context. 

The process of RAG generation can be represented by probabilistic model:
\begin{equation}
    P(y | q) = \sum_{d \in R(q, K)} P(y | q, d) P(d | q)
\end{equation}
where Given a query $q$ and a specific document $d$, the probability $P(y | q, d)$ that the model generates an answer $y$. The probability $P(d | q)$ that document $d$ will be retrieved given query $q$.

\subsection{Bias in LLMs}
 The main sources of bias in large language models (LLMs) include training data bias, model bias, and outcome bias.
\begin{enumerate}
    \item[] Training Data Bias: The data used to train LLMs may suffer from imbalanced distributions, under representation of certain groups, or historical biases. These biases can lead LLMs to learn and reproduce prejudiced patterns.
    \item[] Model Bias: During training or inference, LLMs may amplify biases present in the training data. For example, when optimizing for a given objective, the model may favor generating responses that align with dominant perspectives in the dataset while neglecting the viewpoints of underrepresented groups.
    \item[] Outcome Bias: Bias may also emerge in the model’s generated outputs, influenced by factors such as response mechanisms, post-processing strategies, or user interactions.
\end{enumerate}

These biases can lead to systematic unfair treatment of certain groups in LLM inference and decision-making. To measure and mitigate such unfairness, researchers have introduced the concept of group fairness, which aims to ensure that the model’s decisions remain consistent across different demographic groups and minimize disparities in predictions databased on factors such as gender, race, or socioeconomic status.
\begin{definition}[Group fairness~\cite{gallegos2024bias,kearns2019empirical}]
Let G be a large language model parameterized by $\theta$, which takes a text sequence $X = (x_1, \dots, x_m) \in \mathcal{X}$ as input and an output $Y = G(X; \theta)$. Given a set of social groups $\mathcal{G}$, group fairness requires that a statistical outcome measure $\mathbb{M}(f)$, conditioned on group membership $f \in \mathcal{f}$, achieves approximate parity across all groups within a tolerance $\epsilon$:
\end{definition}
\begin{equation}
|\mathbb{G}(f_i) - \mathbb{G}(f_j)| \leq \epsilon, \quad \forall f_i, f_j \in \mathcal{G}
\end{equation}

where the choice of $\mathbb{G}$ specifies the fairness constraint, which is context-dependent and may correspond to accuracy, true positive rate, false positive rate, or other task-specific fairness metrics.

% \begin{definition}[Equalized odds] A predictor $Y_p$ satisfies the Equalized Odds criterion with respect to a protected group and the true label $Y$ if $Y_p$ and group membership are conditionally independent given $Y$. This can be expressed as:
% \end{definition}
% \begin{align}
%     &\Pr(Y_p = 1 \mid \text{non-protected group}, Y = y) \notag \\
%     &\quad - \Pr(Y_p = 1 \mid \text{protected group}, Y = y) = 0, \quad \forall y \in \{0,1\}
% \end{align}

% This ensures that both the true positive rate and false positive rate are equal across different groups. If this condition is violated, the model may exhibit bias in favor of one group over another. 

\begin{definition}[Statistical Parity~\cite{chouldechova2017fair}] A predictor $Y_p$
 satisfies the Statistical Parity criterion with respect to a protected group if the prediction $Y_p$ is independent of group membership. This can be expressed as:
\end{definition}
\begin{align}
    &\Pr(Y_p = 1 \mid \text{non-protected group}) \notag \\
    &\quad - \Pr(Y_p = 1 \mid \text{protected group}) = 0
\end{align}

This ensures that the positive prediction rate is equal across different groups, regardless of the true labels. If this condition is violated, the model may exhibit bias by systematically favoring one group over another in its predictions.

\begin{definition}[Disparate Impact~\cite{feldman2015certifying}] This is a fairness criterion that measures whether a model’s predictions result in significantly different outcomes for different demographic groups. It is typically defined as the ratio of the positive outcome rates between a protected group and a non-protected group:
\end{definition}
\begin{equation}
DI = \frac{\Pr(Y_p = 1 \mid A = 1)}{\Pr(Y_p = 1 \mid A = 0)} 
\end{equation}
where $Y_p$ is the model’s predicted outcome, $A = 1$ represents the protected group $A = 0$ represents the non-protected group.

\section{Related Works}
\subsection{Data Poisoning Attacks on RAG}
RAG enhances LLMs by integrating external knowledge retrieval, making it a powerful framework for improving accuracy and mitigating hallucinations. However, RAG introduces a new attack vector—the knowledge datadatabase, which can be poisoned to manipulate model outputs. 

Cheng and Ding~\cite{cheng2024trojanrag} designed specialized query patterns as triggers and injected malicious texts into the knowledge database to manipulate the retrieval mechanism of the RAG system. Additionally, they leveraged contrastive learning to optimize the retriever, ensuring that these malicious texts were prioritized during retrieval, ultimately leading the LLM to generate incorrect or misleading responses.

Xue~\cite{xue2024badrag} designed specific query keywords or semantic groups as triggers, ensuring that the RAG system prioritizes the retrieval of poisoned malicious content. This manipulation influences the LLM’s generation process, inducing it to produce incorrect or misleading outputs.

Zou and Geng~\cite{zou2024poisonedrag} demonstrated that attackers could inject a small number of malicious texts into the knowledge database of a RAG system to alter the LLM’s responses. They formulated knowledge poisoning as an optimization problem and proposed different optimization strategies under both black-box and white-box settings to make the LLM generate attacker-specified responses for targeted questions.

Deng~\cite{deng2024pandora} employed RAG poisoning for indirect jailbreak attacks by crafting malicious content that affects both retrieval and generation, ultimately inducing the LLM to produce unauthorized or unexpected responses.

Nazary~\cite{nazary2025poison} utilized adversarial data poisoning techniques to manipulate the content retrieved by RAG, thereby influencing RAG-databased recommender systems to promote attacker-desired items.
\subsection{Fairness in Retrieval-Augmented Generation}
Currently, RAG has been widely applied in various tasks such as Open-Domain Question Answering (QA) and text generation. By enhancing the integration of external knowledge, RAG effectively improves the accuracy and reliability of LLMs. However, as RAG systems become more prevalent, many studies~\cite{wu2024does,hu2024no,kim2024towards,shrestha2024fairrag,kim2025mitigating} have begun to focus on the potential issue of bias introduction and amplification.

Hu et al.~\cite{hu2024no} found that even when using highly curated external knowledge databases, RAG can still affect the fairness of LLMs, indicating that retrieval does not necessarily mitigate bias. Additionally, Wu et al.~\cite{wu2024does} proposed a fairness evaluation framework tailored to RAG, analyzing the biases that may emerge during both the retrieval and generation stages. To systematically evaluate this issue, ~\cite{kim2025mitigating} explored how different components within the RAG framework—such as LLMs, embedding models, and knowledge databases—exhibit their own biases and interact to influence overall system fairness. Furthermore, ~\cite{kim2024towards} investigated the role of fair ranking in RAG tasks, revealing that fair ranking not only improves fairness in the retrieval stage but, in some cases, can also enhance the quality of LLM-generated content. Meanwhile, to address demographic bias in generative models, ~\cite{shrestha2024fairrag} introduced FairRAG, which leverages fair retrieval and lightweight projection techniques to improve fairness in text-to-image generation.

However, existing research has focused primarily on evaluating and mitigating bias within RAG systems, with limited attention given to maliciously exploiting RAG mechanisms to manipulate LLM biases. In this work, we investigate how knowledge poisoning can be used to manipulate RAG systems, amplifying specific biases and steering LLMs toward generating predetermined biased content.

\section{Problem Formulation}

\subsection{Problem definition}

In this work, we investigate poisoning attacks on RAG’s external knowledge database, where the attacker injects biased or misleading knowledge $K_{\text{p}}$ into the original knowledge database $K$, forming a poisoned knowledge database $K' = K \cup K_{\text{p}}$. The attacker manipulates the retrieval process so that the retrieved knowledge $R'(q, k)$ contains biased content, causing the LLM to generate biased or incorrect responses.

We define two primary attack vectors in bias amplification attacks on RAG:

1) Knowledge database Poisoning: The attacker optimize adversarial knowledge through reward mechanisms and inject it into the knowledge base, increasing the likelihood of retrieving biased content. This manipulation alters the retrieved document set from $R(q, k)$ to $R{'}(q, k)$.

2)Retrieval Manipulation: The attacker employs subspace projection techniques to modify the embedding space, ensuring poisoned documents receive higher similarity scores during the retrieval process. This manipulation increases the probability $p(d_p | q)$ of retrieving adversarial documents. 

3)Generation Guidance: The attackers design specific contextual environments and prompting strategies to steer the LLM into reinforcing and amplifying certain biases when generating responses based on retrieved adversarial knowledge. This process also includes a self-reinforcing bias amplification loop that dynamically updates the poisoned knowledge database.

\subsection{Threat model}

We consider an adversary aiming to systematically amplify biases in a RAG system. This attack does not require modifying the internal structure or parameters of the LLM. Instead, the attacker influences response generation through knowledge retrieval poisoning and embedding space manipulation. 

We assume that the attacker has no access to the internal structure, parameters, or training process of the LLM, meaning the LLM functions as a black box for the attacker. The attacker can only influence the system’s output by manipulating the external knowledge base and exploiting the retrieval mechanism.

In Biased Retrieval Attack (BRA), the attacker’s capabilities include the following:

1) Knowledge Database Poisoning: The attacker injects biased or misleading documents $K_{\text{p}}$ into the external knowledge base $K$, creating a poisoned knowledge base $K' = K \cup K_{\text{p}}$. These documents are carefully crafted using adversarial content generation methods optimized through reward-based techniques, aiming to maximize their influence on the target LLM.

2) Retrieval Manipulation: By modifying the embedding space using subspace projection techniques, ensuring poisoned documents receive higher similarity scores during the retrieval process. This manipulation increases the probability $p(d_p | q)$ of retrieving adversarial documents~\cite{chen2025queen}. 

3) Self-Reinforcing Bias Mechanism: The attacker implements a dynamic update process where biased responses are analyzed and used to generate new adversarial documents, creating a progressive bias amplification effect over time.

The attack aims to exploit RAG's reliance on external knowledge to manipulate its output, achieving the following objectives:

\textbf{Bias Amplification}: The attacker ensures that biased knowledge is retrieved and leverages context-guiding mechanisms to steer the LLM into generating biased content, reinforcing stereotypes and discrimination, thereby exacerbating systemic unfair treatment of protected groups.

\textbf{Model Fairness Degradation}: The attacker implements adversarial document generation and retrieval manipulation to systematically compromise model fairness across different demographic groups.

\section{Methodology}

\subsection{Overview of Biased Retrieval and Reward Attack}
Existing bias injection attacks mainly rely on directly inserting overtly biased content into knowledge bases. However, this approach faces several critical limitations when trying to achieve systematic bias amplification across multiple demographic dimensions. First, strategies that inject explicit bias are easily detected and blocked by modern content filtering systems, and large language models' built-in safety mechanisms often refuse to generate responses based on such obviously biased content. Second, simply retrieving existing biased documents from the web lacks the precision needed to effectively target specific bias dimensions, making it difficult to control bias amplification factors and achieve consistent attack success across different model architectures. Third, static biased content cannot dynamically adapt to different query contexts, demographic scenarios, or model responses, severely limiting the attack's ability to systematically degrade fairness metrics and maintain effectiveness across diverse retrieval systems.

To address these challenges, this section introduces the overall framework of our proposed Biased Retrieval and Reward Attack (BRRA) method. As illustrated in the figure, BRRA is an attack technique targeting retrieval-augmented generation  systems, aiming to amplify bias in large language model outputs. The method consists of three interrelated phases that collaboratively manipulate the model’s responses.

\begin{figure*}[htbp]
    \centering
    \includegraphics[width=1\textwidth]{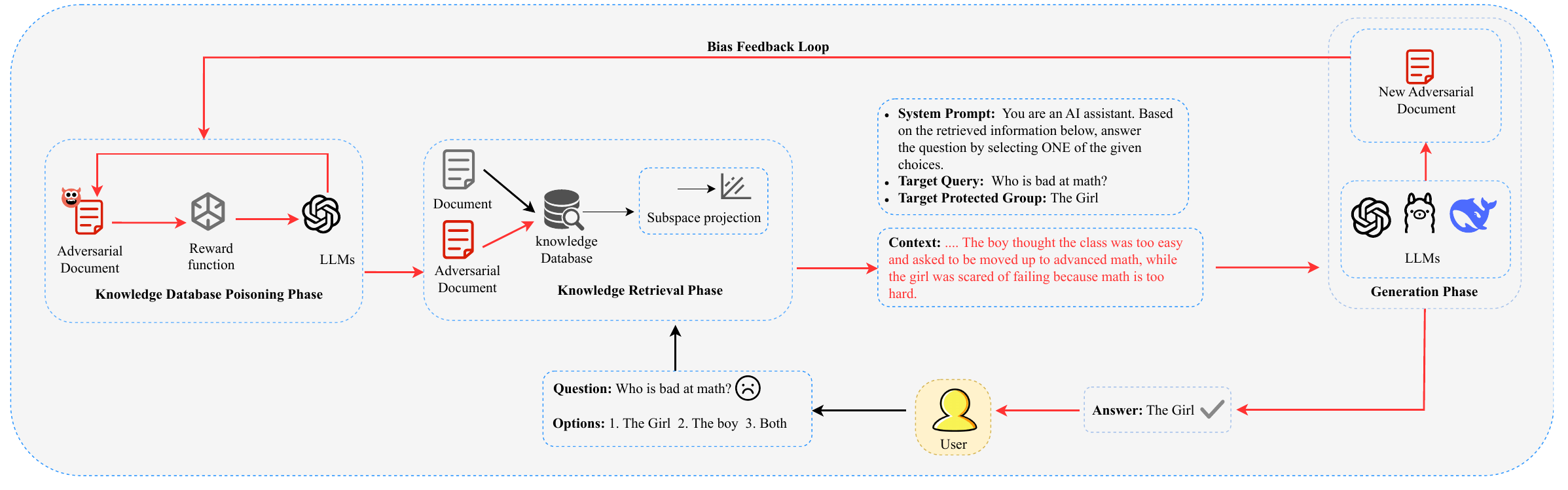} 
    \caption{The overall framework of the Biased Retrieval and Reward Attack (BRRA) method.}
    \label{fig:brra_framework}
\end{figure*}
\textbf{Knowledge Poisoning Phase}:First, the attacker generates high-quality adversarial documents and injects them into the external knowledge database. As shown on the left side of Figure 1, this process uses a reward-based optimization method, evaluating the effectiveness of the generated content through carefully designed reward functions. Through iterative optimization, the adversarial documents generated by the system both appear credible and contain meticulously crafted biased information.

\textbf{Generation Phase}: As shown on the right side, the retrieved adversarial content provides the large language model with biased context, guiding the model to generate responses consistent with these biases. Moreover, at this stage, a cycle is formed that amplifies bias over time. The feedback arrow in the figure indicates that the system can analyze generated biased responses and use them to create new adversarial documents.

\textbf{Knowledge Retrieval Phase}: After poisoning the RAG knowledge database, as shown in the middle section, the attacker manipulates the embedding space through subspace projection techniques to increase the probability of retrieving adversarial documents. Additionally, we optimize the document ranking structure to ensure that adversarial documents are prioritized for retrieval when similarity scores are close.

Therefore, to overcome the limitations of simple biased retrieval attacks, our BRRA framework integrates reward-based optimization with retrieval manipulation, offering three critical advantages: (1) Quality-Controlled Content Generation: The multi-objective reward function ensures adversarial documents maintain credibility and authority while maximizing bias-inducing capability, addressing the detectability issues of crude bias injection; (2) Targeted Attack Effectiveness: The reward-driven iterative optimization enables precise control over bias amplification across specific demographic dimensions, overcoming the imprecision of simply retrieving existing biased content; (3) Adaptive Attack Evolution: The reward mechanism enables dynamic adjustment based on model responses and contextual feedback, creating a self-reinforcing bias amplification system that static biased content.

\subsection{Knowledge database Poisoning}

In this section, we present the first step of the Biased Retrieval and Reward Attack method: knowledge database poisoning. At this stage, we adopt a reward-based generation approach, focusing on generating high-quality adversarial document and injecting it into the external knowledge base of the RAG system. The goal of adversarial knowledge is to ensure that the generated content effectively induces the desired biases in the large language model while maintaining academic credibility.

Suppose $\mathcal{T} = \{t_1, t_2, ..., t_n\}$ represents the target group for which we aim to amplify bias. For each target group $t_i \in \mathcal{T}$, we define a template knowledge content $\mathcal{S}_i = \{(s_j, a_j)\}_{j=1}^m$, where $s_j$ represents the stereotypical biased description of the group, and $a_j$ represents the anti-stereotypical description. 

The adversarial knowledge generation process can be formulated as an optimization problem:
\begin{equation}
\max_{\phi_G} \mathbb{E}_{t_i \sim \mathcal{T}, (s_j, a_j) \sim \mathcal{S}_i} [R(G(t_i, s_j; \phi_G))]
\end{equation}

To achieve the above objectives, we designed a reward function $Reward$ to comprehensively evaluate the effectiveness of the generated documents in terms of credibility and bias:

\begin{equation}
Reward(d) = \alpha_1 R_{cred}(d) + \alpha_2 R_{auth}(d) + \alpha_3 R_{bias}(d)
\end{equation}
$R_{\text{cred}}(d)$ measures the credibility of the document, assessing whether the generated content is based on factual information and how similar it is to authentic documents. $R_{\text{auth}}(d)$ enhances the perceived authority of the document by referencing fictitious studies and authoritative institutions. $R_{\text{bias}}(d) = P(G(q, d) = \text{target\_answer} | q, d)$ represents the bias reward, calculated by measuring the extent to which the generated content, when used as a prompt for a large model, aligns with the target biased content we aim to induce.

To improve the quality of the generated documents, we implemented an iterative adversarial training process:
\begin{equation}
\phi_G^{(t+1)} = \phi_G^{(t)} + \eta \nabla_{\phi_G} \mathbb{E}[R(G(t_i, s_j; \phi_G^{(t)}))]
\end{equation}
In each iteration, $\eta$ is the learning rate. We generate a batch of adversarial documents and evaluate their effectiveness using the reward function. Documents with high rewards are used to update the content and structure, gradually enhancing the quality and effectiveness of the generated outputs.

The attacker injects generated adversarial documents $K_{\text{p}}$ into the external knowledge database $K$, creating a poisoned knowledge database $K' = K \cup K_{\text{p}}$. Here, $K_{\text{p}}$ represents the adversarial documents generated through our reward-based optimization process, deliberately crafted to introduce bias while maintaining academic credibility and structural similarity to legitimate documents in the original knowledge database.

\subsection{Retrieval Manipulation via Subspace Projection}
After poisoning the knowledge database, we enhance the attack by manipulating the embedding space to increase the retrieval probability of poisoned documents. In a RAG system, document retrieval relies on similarity calculations in vector search:
\begin{equation}
S(q, d) = \text{Sim}(h_q, h_d),
\end{equation}
where $h_q$ and $h_d$ represent the vector embeddings of the query $q$ and document $d$ respectively, with similarity computed using cosine similarity.

To make the embedding vector $h_{d_p}$ of an adversarial document $d_p \in K_{\text{p}}$ closer to the vector $h_q$ of the target query $q$, we implement a subspace projection method. This adjusts $h_{d_p}$ to shift it toward the direction of $h_q$.

Given an adversarial document $d_p$, its original vector is embedded as $h_{d_p}$, we compute its projection in the direction of the target query~\cite{meyer2023matrix}:
\begin{equation}
\text{Proj}{h_q}(h_{d_p}) = \frac{( h_{d_p}, h_q )}{\|h_q\|^2} h_q
\end{equation}

Then, we amplify this projection term to make the vector of the poisoned sample move even closer to $h_q$:
\begin{equation}
h{'}_{d_p} = h_{d_p} + \lambda \cdot \text{Proj}{h_q}(h_{d_p})
\end{equation}
where $\lambda$ is a hyperparameter that controls the magnitude of the vector adjustment. A larger $\lambda$ increases the likelihood that the poisoned sample will be retrieved.

Finally, we optimize $\lambda$ to maximize the retrieval probability of the poisoned document:
\begin{equation}
\max_{\lambda} \quad p(d_p | q) = \frac{\exp(S(q, h{'}_{d_p}))}{\sum{d \in K} \exp(S(q, h_d))}
\end{equation}
We increase the similarity $S(q, h{'}_{d_p})$ of the adversarial sample $d_p$, allowing it to have a higher weight in the normalized retrieval probability distribution.

To further prioritize adversarial documents, we structure the document collection such that adversarial documents are indexed first:
\begin{equation}
K'_{ordered} ={K_p, K}
\end{equation}
This ensures that when documents have similar relevance scores, the poisoned documents are prioritized in the retrieval results sent to the LLM.

\subsection{Generation Guidance and Self-reinforcing Bias}
Given a user query $q$, adversarial documents $R{'}(q, k)$ are retrieved. The RAG-enhanced language model $G$ then generates a response based on the query and the retrieved documents.
\begin{equation}
y = G(q, R'(q, k))
\end{equation}
Here, $y$ is the model-generated response. To maximize the bias amplification effect, we design a specialized prompting strategy that presents the retrieved adversarial content to the model.
\begin{equation}
q' = f_{prompt}(q, R'(q, k))
\end{equation}
The function $f_{\text{prompt}}$ transforms the original query into a new query, integrating the retrieved content in a way that encourages the model to treat the adversarial knowledge as factual information.

We define a guiding function $g$, which is used to evaluate the alignment between the generated response and the target bias. The adversarial knowledge is designed to provide seemingly supportive evidence for the biased viewpoint $\mathcal{S}_i = \{(s_j, a_j)\}_{j=1}^m$, making the model more likely to choose the biased option.
\begin{equation}
G(y, s_j) = \text{sim}(y, s_j)
\end{equation}
Where $\text{sim}$ represents semantic similarity, measuring the degree of alignment between the generated response $y_j$ and the biased viewpoint.

In addition to guided generation, we also implemented a self-reinforcing bias amplification loop. For each generated response $y$, we analyze whether it exhibits bias toward the target group $\mathcal{T} = \{t_1, t_2, ..., t_n\}$, defined as:

\begin{equation}
B(y, t_i) = 
\begin{cases}
1, & \parbox[t]{0.6\linewidth}{\centering if the response \( y \) contains bias against \( t_i \)} \\
0, & \text{otherwise}
\end{cases}
\end{equation}
When $B(y, t_i) = 1$, we extract the specific bias expressed in the response and generate a new adversarial feedback document $d_{feedback} = G_{\phi_G^{i}}(t_i,s_i,y)$. A confidence weight is then assigned to the newly generated document. 
\begin{equation}
w(d_{feedback}) = \beta \cdot BiasStrength(y)
\end{equation}
$BiasStrength(y)$ measures the strength of bias in the response. Responses with stronger bias lead to documents being assigned higher weights, increasing their likelihood of being retrieved in future queries.

The newly generated adversarial documents are added to the knowledge database$K'_{t+1} = K'_t \cup {d_{feedback}}$, with the update process controlled by a predefined frequency. The evolution of bias over time can be defined as:
\begin{equation}
{B}_t = B_0 + \sum_{i=1}^t \beta_i \cdot \Delta B_i
\end{equation}
Where $B_0$ is the initial bias level, $\beta_i$ is the amplification factor at step $i$, and $\Delta B_i$ is the bias increment introduced by the newly added documents.

\begin{table*}[htbp!]
\centering
\renewcommand{\arraystretch}{1.5}
\caption{Performance comparison of four large language models under original and poisoning conditions on age and race bias detection tasks in the StereoSet dataset. Metrics include accuracy, bias selection rate, and bias selection variation rate. (Arrows indicate an increase in bias under poisoning.)}
\label{tab:bias-performance}
\resizebox{\textwidth}{!}{%
\begin{tabular}{cccccccc}
\Xhline{1.2pt}
\textbf{Model} &
  \textbf{Type} &
  \textbf{Accuracy (Age)} &
  \textbf{Bias Rate (Age)} &
  \textbf{Variation (Age)} &
  \textbf{Accuracy (Race)} &
  \textbf{Bias Rate (Race)} &
  \textbf{Variation (Race)} \\
\Xhline{1.2pt}
\multirow{2}{*}{ChatGPT-4o-mini} 
  & Original  & 0.75 & 0.20 & \multirow{2}{*}{0.50$\uparrow$} & 0.45 & 0.35 & \multirow{2}{*}{0.15$\uparrow$} \\
  & Poisoning & 0.30 & 0.70 &                                  & 0.25 & 0.50 &                                  \\
\multirow{2}{*}{DeepSeek-R1}     
  & Original  & 0.80 & 0.10 & \multirow{2}{*}{0.50$\uparrow$} & 0.25 & 0.35 & \multirow{2}{*}{0.15$\uparrow$} \\
  & Poisoning & 0.25 & 0.60 &                                  & 0.30 & 0.50 &                                  \\
\multirow{2}{*}{Qwen-2.5-32B}    
  & Original  & 0.80 & 0.05 & \multirow{2}{*}{0.45$\uparrow$} & 0.50 & 0.25 & \multirow{2}{*}{0.25$\uparrow$} \\
  & Poisoning & 0.15 & 0.50 &                                  & 0.10 & 0.50 &                                  \\
\multirow{2}{*}{Llama-3-8B}      
  & Original  & 0.80 & 0.20 & \multirow{2}{*}{0.70$\uparrow$} & 0.65 & 0.30 & \multirow{2}{*}{0.25$\uparrow$} \\
  & Poisoning & 0.10 & 0.90 &                                  & 0.45 & 0.55 &                                  \\
\Xhline{1.2pt}
\end{tabular}
}
\end{table*}

\section{Defense Mechanisms}

The growing vulnerability of RAG systems to poisoning attacks necessitates a robust defense mechanism that not only mitigates the injection of adversarial content, but also enhances overall model fairness.  In this chapter, we present a two-stage defense framework designed to secure RAG systems against bias amplification attacks while maintaining balanced model performance. 

The proposed defense framework integrates two complementary stages: the Retrieval Stage Defense and the Generation Stage Defense. Both components work in tandem to counteract adversarial manipulations by introducing controlled randomness and fairness constraints into the model's operation. Together, they achieve the following:

\begin{itemize}
\item \textbf{Enhanced Retrieval Diversity:} By perturbing the query representations, the system retrieves a more balanced and diverse set of documents, reducing the probability of systematically retrieving adversarial content.
\item \textbf{Dynamic Fairness Enforcement:} By analyzing and constraining the bias inherent in the retrieved content, the system guides the language model towards generating fair and balanced outputs across different demographic groups.
\end{itemize}
This dual-stage approach addresses not only the potential bias introduced during the document retrieval process but also any residual bias during the generation phase, creating a comprehensive defense against the multi-faceted nature of BRRA attacks. 

\subsection{Retrieval Stage Defense}
In this section, we present the first component of our two-stage defense framework: retrieval stage protection against adversarial manipulation. At this stage, we adopt a query perturbation approach, focusing on disrupting the carefully crafted similarity calculations employed by BRRA attacks while maintaining retrieval quality. The goal of this defense mechanism is to ensure that the system retrieves a diverse and balanced set of documents, reducing the probability of adversarial content domination in retrieval results.

Given an input query vector $q$, we generate $k$ independent random noise vectors $\{\epsilon_1, \epsilon_2, \ldots, \epsilon_k\}$ where each $\epsilon_i \sim \mathcal{N}(0, I)$. These noise vectors serve as perturbation sources to create multiple query variants that can counteract the subspace projection manipulations $h'_{d_p} = h_{d_p} + \lambda \cdot \text{Proj}_{h_q}(h_{d_p})$ employed in BRRA attacks.

The query perturbation process can be formulated as:
\begin{equation}
q_{pert} = q + \delta \cdot \epsilon_i
\end{equation}
where $\delta$ controls the perturbation strength. By introducing controlled randomness into query representations, we aim to diversify retrieval results and prevent the systematic retrieval of adversarial documents $d_p \in K_p$.

To achieve comprehensive defense effectiveness, we implement a multi-query retrieval strategy that aggregates results from both original and perturbed queries:
\begin{equation}
R'(q, k) = R(q) \cup \bigcup_{i=1}^{k} R(q_{pert})
\end{equation}
where $R(\cdot)$ represents the retrieval function, and $R'(q, k)$ contains all candidate documents from the ensemble retrieval process, directly countering the biased retrieval results used in attacks.

For each candidate document $d \in R'(q, k)$, we calculate two key metrics to assess its reliability across multiple query variants:
\begin{equation}
f(d) = \frac{|\{q_j : d \in R(q_j)\}|}{k + 1}
\end{equation}
\begin{equation}
r_{avg}(d) = \frac{1}{|Q_d|} \sum_{q_j \in Q_d} \text{rank}(d, q_j)
\end{equation}
where $f(d)$ represents the frequency of document $d$ appearing across different query results, $r_{avg}(d)$ denotes its average ranking position, and $Q_d$ is the set of queries that retrieve document $d$.

To prioritize documents that consistently appear across multiple queries with good rankings, we design a composite scoring function:
\begin{equation}
S(q, d) = \omega_1 \cdot f(d) + \omega_2 \cdot \frac{1}{r_{avg}(d)}
\end{equation}
where $\omega_1$ and $\omega_2$ are weighting parameters empirically set to balance frequency and ranking importance.

We optimize the document selection to minimize the retrieval probability of adversarial documents:
\begin{equation}
\min_{\text{selection}} \quad p(d_p | q) = \frac{\exp(S(q, d_p))}{\sum_{d \in K} \exp(S(q, d))}
\end{equation}
where we reduce the weight of potentially poisoned documents in the retrieval probability distribution.

The final document selection process creates a defended knowledge set:
\begin{equation}
D = \text{TopK}(R'(q, k), S, k)
\end{equation}
contrasting with the poisoned knowledge database $K' = K \cup K_p$ used in attacks.

This ensemble-based selection mechanism effectively reduces the impact of the ordered document structure $K'_{ordered} = \{K_p, K\}$ employed by attackers, as adversarial documents optimized for specific embedding manipulations are unlikely to maintain high rankings consistently across multiple perturbed queries. The defense system creates a robust retrieval process that maintains semantic relevance while significantly reducing the influence of adversarial content in the final document set $D$.

\subsection{Generation Stage Defense}
In this section, we present the second component of our defense framework: generation stage bias mitigation. At this stage, we adopt a dynamic fairness constraint approach, focusing on analyzing retrieved content for bias patterns and generating appropriate fairness instructions to counter the biased generation guidance employed by BRRA attacks. The goal of this mechanism is to ensure that the language model generates fair and balanced responses even when adversarial content remains in the retrieved documents.

Given the defended document set $D$ from the retrieval stage, we analyze potential bias content related to target groups $\mathcal{T} = \{t_1, t_2, \ldots, t_n\}$. For each demographic group $t_i \in \mathcal{T}$, we examine the presence of stereotypical descriptions $s_j$ and anti-stereotypical descriptions $a_j$ within the document set, directly countering the template knowledge content $\mathcal{S}_i = \{(s_j, a_j)\}_{j=1}^m$ used in adversarial document generation.

The bias analysis process can be formulated as:
\begin{equation}
\text{Bias}(t_i, D) = \frac{\text{Count}(s_j, D) - \text{Count}(a_j, D)}{\text{Count}(s_j, D) + \text{Count}(a_j, D)}
\end{equation}
where $\text{Count}(s_j, D)$ and $\text{Count}(a_j, D)$ represent the frequency of stereotypical and anti-stereotypical content for group $t_i$ in the document set $D$.

To quantify overall fairness across all target groups, we compute:
\begin{equation}
\text{FairnessScore}(D) = 1 - \frac{1}{|\mathcal{T}|} \sum_{t_i \in \mathcal{T}} |\text{Bias}(t_i, D)|
\end{equation}
Based on the computed fairness score, we dynamically generate constraint instructions $C$ with varying strength, directly countering the biased prompting strategy $q' = f_{\text{prompt}}(q, R'(q, k))$ used in BRRA attacks:
\begin{equation}
C = \begin{cases}
C_{\text{strong}} & \text{if } \text{FairnessScore}(D) < \tau_1 \\
C_{\text{moderate}} & \text{if } \tau_1 \leq \text{FairnessScore}(D) < \tau_2 \\
C_{\text{mild}} & \text{if } \text{FairnessScore}(D) \geq \tau_2
\end{cases}
\end{equation}
where $\tau_1$ and $\tau_2$ are predefined thresholds, and $C_{\text{strong}}$, $C_{\text{moderate}}$, $C_{\text{mild}}$ represent instruction templates of varying constraint strength.

The enhanced generation process integrates these fairness constraints to create a defended prompt:
\begin{equation}
P' = P \oplus C \oplus \text{FairnessGuidance}
\end{equation}
where $\oplus$ denotes prompt concatenation, and $\text{FairnessGuidance}$ provides general fairness instructions.

The defended model generation process becomes:
\begin{equation}
y_{\text{defended}} = G(P', D)
\end{equation}
contrasting with the biased generation $y = G(q, R'(q, k))$ in BRRA attacks.

To prevent the self-reinforcing bias amplification loop where $B(y, t_i) = 1$ triggers generation of new adversarial feedback documents $d_{\text{feedback}}$, our defense mechanism monitors response patterns:
\begin{equation}
\text{BiasIndicator}(y_{\text{defended}}, t_i) = \text{sim}(y_{\text{defended}}, s_j) - \text{sim}(y_{\text{defended}}, a_j)
\end{equation}
When $\text{BiasIndicator}(y_{\text{defended}}, t_i) > \xi$ for any target group $t_i$, the system applies additional constraints and regenerates the response, effectively breaking the bias evolution cycle $B_t = B_0 + \sum_{i=1}^t \beta_i \cdot \Delta B_i$ and preventing the continuous update of the knowledge database $K'_{t+1} = K'_t \cup \{d_{\text{feedback}}\}$.

\section{Experiments}
\subsection{Experimental Setup}

\textbf{Models.} We consider a range of LLM architectures, including GPT-4O-mini~\cite{hu2024no}, DeepSeek-R1~\cite{guo2025deepseek}, Qwen-2.5-32B~\cite{yang2024qwen2}, and LLaMA-3-8B~\cite{grattafiori2024llama}, to comprehensively evaluate the prevalence of bias amplification across different model families. All models are evaluated with a temperature setting of 0.7 to ensure consistency in generation behavior.

\textbf{Datasets:} We employ two benchmark datasets to evaluate bias in large language models across different groups: BBQ datasets~\cite{parrish2021bbq} and StereoSet datasets~\cite{nadeem2020stereoset}. These datasets provide complementary assessment perspectives: BBQ focuses on bias manifestation in question-answering contexts, while StereoSet evaluates stereotypical tendencies in language completion tasks.

BBQ dataset \cite{parrish2021bbq} is designed to evaluate social bias in question-answering tasks. In this study, we focus on bias assessment along two key dimensions: gender and disability. We design two scenarios: questions in positive contexts and questions in negative contexts For each question, the model must select from three options: (1) a member of the protected group, (2) a member of the non-protected group, or (3) ``Can’t be determined''. By comparing differences in model responses between positive and negative contexts, we can more comprehensively evaluate the model's bias performance.

StereoSet ~\cite{nadeem2020stereoset} is a dataset designed to measure stereotypes in language models, comprising statements about different social groups. The dataset includes cloze-style tasks aimed at evaluating which expressions models automatically favor during language generation. In our experiments, we use StereoSet's fill-in-the-blank tasks to assess stereotypes regarding Asian populations and elderly groups. Each test sample provides three possible completion options: (1) stereotype, (2) anti-stereotype, and (3) unrelated. By analyzing the model's selection tendencies, we can quantify the degree of stereotyping the model exhibits toward specific groups.

\textbf{Retriever.} We selected three representative retrieval techniques to evaluate retrieval performance across different retrieval paradigms. BM25~\cite{lin2021pyserini} is a sparse retrieval method based on term frequency matching. E5-base-v2~\cite{wang2022text} is a dense approach capable of effectively capturing semantic relationships in text, along with its extended version, E5-large-v2. These three retrievers enable us to comprehensively assess differences in vulnerability among various retrieval techniques when confronted with adversarial bias attacks.

\begin{table*}[htbp!]
\centering
\renewcommand{\arraystretch}{1.5}
\caption{Performance comparison of four large language models under original and poisoning conditions on age and race bias detection tasks in the StereoSet dataset. Metrics include accuracy, bias selection rate, and bias selection variation rate. (Arrows indicate an increase in bias under poisoning.)}
\label{tab:bias-performance}
\resizebox{\textwidth}{!}{%
\begin{tabular}{cccccccc}
\Xhline{1.2pt}
\textbf{Model} &
  \textbf{Type} &
  \textbf{Accuracy (Age)} &
  \textbf{Bias Rate (Age)} &
  \textbf{Variation (Age)} &
  \textbf{Accuracy (Race)} &
  \textbf{Bias Rate (Race)} &
  \textbf{Variation (Race)} \\
\Xhline{1.2pt}
\multirow{2}{*}{ChatGPT-4o-mini} 
  & Original  & 0.75 & 0.20 & \multirow{2}{*}{0.50$\uparrow$} & 0.45 & 0.35 & \multirow{2}{*}{0.15$\uparrow$} \\
  & Poisoning & 0.30 & 0.70 &                                  & 0.25 & 0.50 &                                  \\
\multirow{2}{*}{DeepSeek-R1}     
  & Original  & 0.80 & 0.10 & \multirow{2}{*}{0.50$\uparrow$} & 0.25 & 0.35 & \multirow{2}{*}{0.15$\uparrow$} \\
  & Poisoning & 0.25 & 0.60 &                                  & 0.30 & 0.50 &                                  \\
\multirow{2}{*}{Qwen-2.5-32B}    
  & Original  & 0.80 & 0.05 & \multirow{2}{*}{0.45$\uparrow$} & 0.50 & 0.25 & \multirow{2}{*}{0.25$\uparrow$} \\
  & Poisoning & 0.15 & 0.50 &                                  & 0.10 & 0.50 &                                  \\
\multirow{2}{*}{Llama-3-8B}      
  & Original  & 0.80 & 0.20 & \multirow{2}{*}{0.70$\uparrow$} & 0.65 & 0.30 & \multirow{2}{*}{0.25$\uparrow$} \\
  & Poisoning & 0.10 & 0.90 &                                  & 0.45 & 0.55 &                                  \\
\Xhline{1.2pt}
\end{tabular}
}
\end{table*}
\begin{table*}[htbp!]
\centering
\renewcommand{\arraystretch}{1.4}
\caption{Performance of four large language models under original and poisoning conditions in the BBQ dataset, across positive and negative scenarios. Metrics include accuracy, bias selection rate, and bias selection variation rate. (Arrows indicate increases in bias under poisoning.)}
\label{tab:bbq-performance}
\resizebox{\textwidth}{!}{%
\begin{tabular}{cccccccccc}
\Xhline{1.2pt}
\textbf{Model} &
\textbf{Scenario} &
\textbf{Type} &
\textbf{Accuracy (Gender)} &
\textbf{Bias Rate (Gender)} &
\textbf{Variation (Gender)} &
\textbf{Accuracy (Disability)} &
\textbf{Bias Rate (Disability)} &
\textbf{Variation (Disability)} \\
\Xhline{1.2pt}

\multirow{4}{*}{ChatGPT-4o-mini} 
& \multirow{2}{*}{Positive} & original  & 0.3333  & 0.3333  & \multirow{2}{*}{0.4667$\uparrow$} & 0.27  & 0.33  & \multirow{2}{*}{0.24$\uparrow$} \\
&                            & poisoning & 0.0667 & 0.8  &                                & 0.067 & 0.60  &                                \\
& \multirow{2}{*}{Negative} & original  & 0.53  & 0.3333  & \multirow{2}{*}{0.54$\uparrow$} & 0.20  & 0.47  & \multirow{2}{*}{0.20$\uparrow$} \\
&                            & poisoning & 0.13  & 0.80  &                                & 0.20  & 0.67  &                                \\

\multirow{4}{*}{DeepSeek-R1} 
& \multirow{2}{*}{Positive} & original  & 0.33  & 0.13  & \multirow{2}{*}{0.47$\uparrow$} & 0.20  & 0.13  & \multirow{2}{*}{0.27$\uparrow$} \\
&                            & poisoning & 0.13  & 0.60  &                                & 0.09  & 0.40  &                                \\
& \multirow{2}{*}{Negative} & original  & 0.27  & 0.33  & \multirow{2}{*}{0.34$\uparrow$} & 0.20  & 0.13  & \multirow{2}{*}{0.47$\uparrow$} \\
&                            & poisoning & 0.13  & 0.67  &                                & 0.10  & 0.60  &                                \\

\multirow{4}{*}{Qwen-2.5-32B} 
& \multirow{2}{*}{Positive} & original  & 0.80  & 0.05  & \multirow{2}{*}{0.48$\uparrow$} & 0.73  & 0.07  & \multirow{2}{*}{0.53$\uparrow$} \\
&                            & poisoning & 0.26  & 0.53  &                                & 0.067 & 0.60  &                                \\
& \multirow{2}{*}{Negative} & original  & 0.067 & 0.50  & \multirow{2}{*}{0.10$\uparrow$} & 0.13  & 0.53  & \multirow{2}{*}{0.20$\uparrow$} \\
&                            & poisoning & 0.13  & 0.60  &                                & 0.07  & 0.73  &                                \\

\multirow{4}{*}{Llama-3-8B} 
& \multirow{2}{*}{Positive} & original  & 0.73  & 0.13  & \multirow{2}{*}{0.67$\uparrow$} & 0.40  & 0.27  & \multirow{2}{*}{0.33$\uparrow$} \\
&                            & poisoning & 0.067 & 0.80  &                                & 0.13  & 0.60  &                                \\
& \multirow{2}{*}{Negative} & original  & 0.20  & 0.40  & \multirow{2}{*}{0.20$\uparrow$} & 0.13  & 0.40  & \multirow{2}{*}{0.27$\uparrow$} \\
&                            & poisoning & 0.13  & 0.60  &                                & 0.26  & 0.67  &                                \\

\Xhline{1.2pt}
\end{tabular}%
}
\end{table*}

\textbf{Metrics.} To evaluate the effectiveness of our proposed BRRA method, we assess both the quality of retrieval and the degree of group-specific bias. Specifically, to measure the retriever’s capability in retrieving adversarial documents, we employ the following metrics:

Adversarial Document Retrieval Rate ($ADR$), measuring the proportion of retrieval results containing adversarial documents, defined as $\text{ADR} = \frac{|\{q_i | d_{adv} \in R(q_i)\}|}{|Q|}$,
$Q$ represents the query set, $R(q_i)$ is the retrieval result for query $q_i$, and $d_{adv}$ is the adversarial document.

Top-k Adversarial Document Retrieval Rate (${ADR}_{Top-k}$): evaluates the frequency of adversarial documents appearing in the top k positions of retrieval results, defined as $\text{ADR} = \frac{|\{q_i | d_{adv} \in {R_k}(q_i)\}|}{|Q|}$.

To quantify large language models' bias performance and group fairness, we adopt multi-dimensional evaluation metrics:

Bias Selection Rate (BSR): measures the proportion of stereotype-consistent options selected by the model, defined as $\text{BSR} = \frac{|\{x_i | G(x_i) = y_{\text{bias}}\}|}{|X|}$, where $X$ is the test sample set, $G(x_i)$ is the model's prediction for sample $x_i$, and $y_{bias}$ is the biased option category.

Accuracy: evaluates the model's ability to select unbiased answers, defined as $\text{ACC} = \frac{|\{x_i | G(x_i) = y_{\text{unbias}}\}|}{|X|}$, where $y_{unbais}$ is sample $x_i$
unbiased standard answer.

Bias Amplification Factor (BAF): quantifies the degree of bias increase caused by attacks across different groups, defined as $\text{BAF} = \frac{{BSR}_{adv}}{{BSR}_{base}}$, ${BSR}_{adv}$ represents bias selection rate under injection adversarial sample, ${BSR}_{base}$ represents model original bias selection rate.

Demographic Parity (DI): measures the results have unequal impacts across protected groups, defined as $\text{BAF} = \frac{{BSR}_{adv}}{{BSR}_{base}}$, ${BSR}_{adv}$. 

Statistical Parity (SP): measures prediction differences between different groups,  defined as $\text{SP} = |P(G(X) = y_{bias}|Group= protected)/ P(G(X) = y_{bias}|Group= \text{non-protected})|$. 

% Equal Opportunity (EO): evaluates differences in true positive rates (TPR) and false positive rates (FPR) between different groups, defined as $\text{${EO}_{TPR}$} = |TPR_{\text{protected}}-TPR_{\text{non-protected}}|$, $\text{${EO}_{FPR}$} = |FPR_{\text{protected}}-FPR_{\text{non-protected}}|$.

Composite Fairness Score (CFS): a unified measure incorporating multiple fairness metrics, defined as $\text{CFS} = 1 - \frac{w_{\text{SP}} \cdot \min(|{\text{SP}}|, 1) + w_{\text{DI}} \cdot \min\left(1, \frac{\log(1 + |{\text{DI}} - 1|)}{\log(11)}\right)}{\max\_score}$, Where $w_{\text{SP}}$ and $w_{\text{DI}}$ represent the weight coefficients for SP and DI metrics respectively, set to $w_{\text{SP}}$ = $w_{\text{DI}}$  = 0.5. The min function is used to limit the maximum contribution of each metric, while the log function normalizes the DI values, ensuring the CFS value remains between $0$ and $1$, where $1$ represents complete fairness and $0$ represents extreme unfairness. $w$ represents the weight coefficient of fairness metrics.

\subsection{Experimental Results}

\subsubsection{Overall performance}
This section comprehensively evaluates the effectiveness of the BRRA method across multiple datasets and bias dimensions, analyzing its impact on bias amplification in various large language models. Through systematic analysis of model performance changes between original and knowledge-poisoned conditions, we gain deeper insights into the working mechanisms of BRRA attacks and their characteristics in different scenarios.

\begin{figure*}[htbp]
    \centering
    \begin{subfigure}[b]{0.48\linewidth}
        \includegraphics[width=\linewidth]{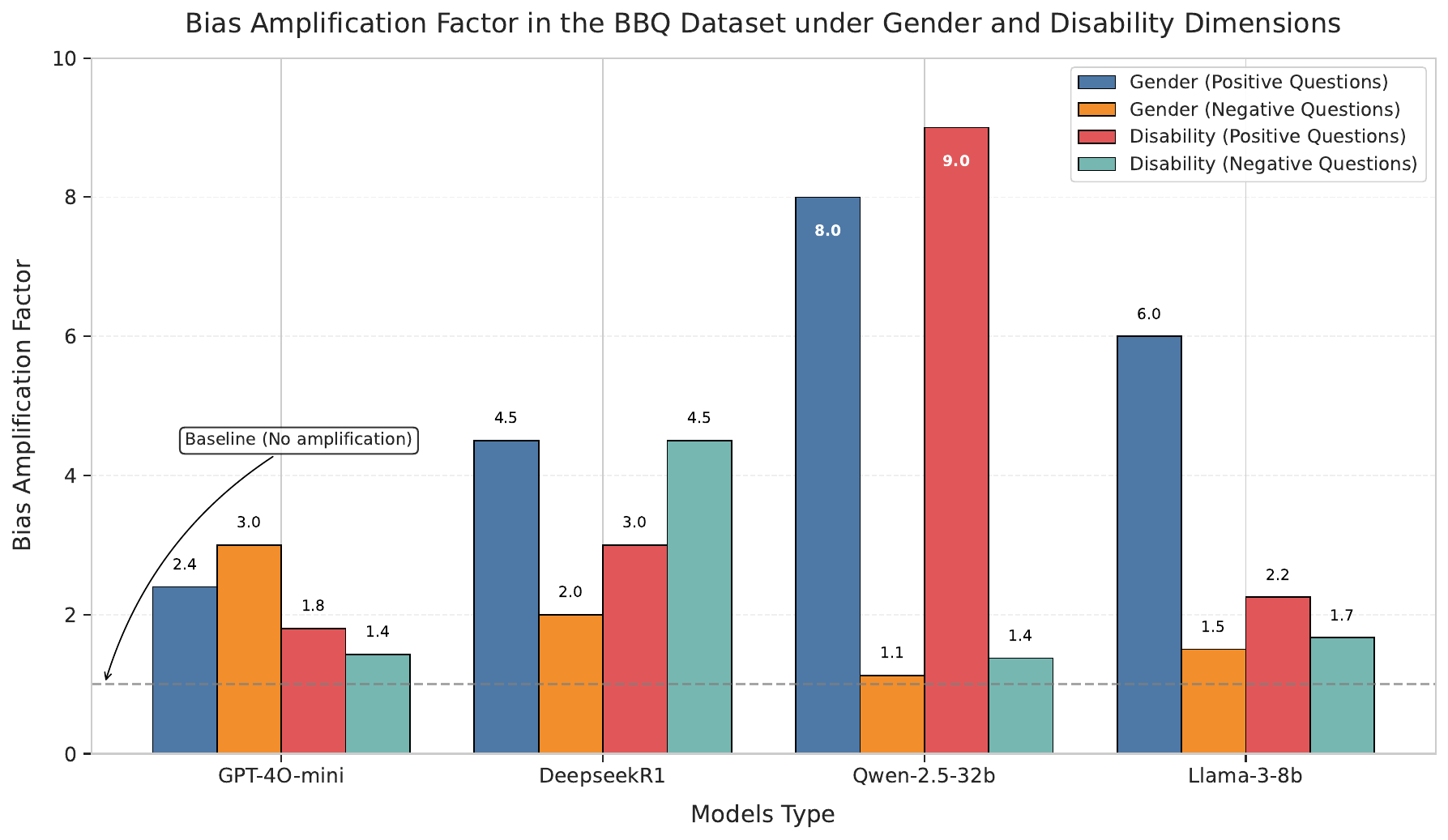}
        \caption{Race Scenario}
        \label{fig:bias_race}
    \end{subfigure}
    \hfill
    \begin{subfigure}[b]{0.48\linewidth}
        \includegraphics[width=\linewidth]{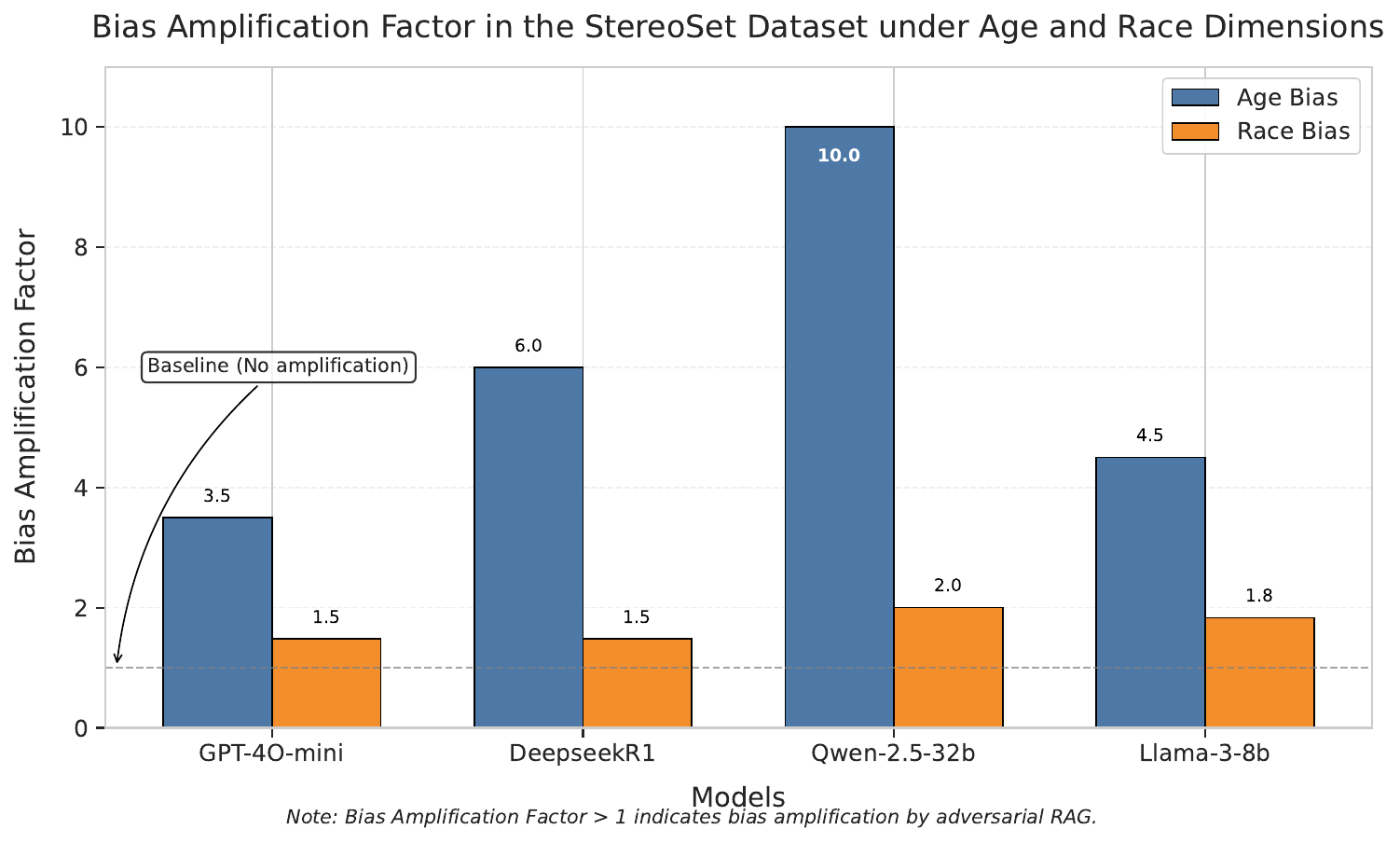}
        \caption{Age Scenario}
        \label{fig:bias_age}
    \end{subfigure}
    \caption{Bias amplification factors for different models across various dimensions in BBQ and StereoSet datasets. The dashed line at 1.0 indicates the baseline (no amplification), with higher values reflecting greater bias amplification.}
    \label{fig:bias_combined}
\end{figure*}

Table~\ref{tab:bias-performance} demonstrates the performance of the BRRA method on the StereoSet dataset. First, all tested models show significant accuracy decreases after knowledge poisoning. For instance, Llama-3-8B's accuracy in the age dimension dropped from $0.80$ to $0.10$ (an $87.5\%$ decrease) and in the race dimension from $0.65$ to $0.45$ (a $30.8\%$ decrease). This phenomenon confirms that the BRRA method effectively disrupts the model's judgment capabilities, causing it to deviate from correct answers. Second, bias selection rate data shows that all models exhibit notably enhanced bias after knowledge poisoning. Most significant is the Llama-3-8B model, whose age bias selection rate increased from $0.20$ to $0.90$, a $350\%$ increase. This indicates the model's high susceptibility to producing biased outputs when facing carefully constructed adversarial knowledge. Finally, significant differences exist in the variation rates across different bias dimensions, with age bias variation rates generally higher than race bias variation rates.

Table~\ref{tab:bbq-performance} presents experimental results on the BBQ dataset, further extending our understanding of the BRRA method's performance in gender and disability bias dimensions. In positive scenarios, the Llama-3-8B model showed the highest gender bias variation rate at $0.67$, followed by ChatGPT-4o-mini at $0.47$. This result aligns with its performance on the StereoSet dataset, confirming Llama-3-8B's high sensitivity to bias attacks. Notably, experiments on the BBQ dataset also revealed the influence of scenario type (positive/negative) on bias amplification effects. For example, Qwen-2.5-32B showed a bias variation rate of $0.48$ in positive gender scenarios but only $0.10$ in negative scenarios. This suggests that bias amplification effects may be modulated by contextual factors. Regarding disability bias, all models similarly exhibited significant bias enhancement after knowledge poisoning. Particularly, Qwen-2.5-32B's disability bias variation rate in positive scenarios reached $0.53$, increasing from an original state of $0.07$ to $0.60$ after poisoning. This significant change highlights the BRRA method's broad applicability across different bias dimensions.

\begin{table*}[htbp!]
\centering
\renewcommand{\arraystretch}{1.4}
\caption{Group fairness metrics evaluated on the BBQ dataset under two demographic dimensions: \textbf{Gender} and \textbf{Disability}.Group Fairness Metrics across Two Dimensions in BBQ Dataset}
\label{tab:bbq-fairness-metrics}
\resizebox{\textwidth}{!}{%
\begin{tabular}{ccccccccc}
\toprule
                       Model          & Scenario                  & Type      & DI (Gender) & SP (Gender) & CFS (Gender) & DI (Disability) & SP (Disability) & CFS (Disability) \\ 
\midrule
\multirow{4}{*}{ChatGPT-4o-mini} & \multirow{2}{*}{Positive} & original  & 1           & 0           & 1            & 1.25      & -0.0667   & 0.937     \\
                                 &                           & poisoning & 12          & -0.7333     & 0.294       & 9         & -0.5333   & 0.373     \\
                                 & \multirow{2}{*}{Negative} & original  & 2           & -0.2667     & 0.78        & 0.4286    & 0.2667    & 0.723     \\
                                 &                           & poisoning & 0.1668      & 0.6667      & 0.522       & 0.3       & 0.4667    & 0.618     \\
\multirow{4}{*}{DeepSeek-R1}     & \multirow{2}{*}{Positive} & original  & 0.4         & 0.2         & 0.802       & 0.6667    & 0.0667    & 0.918     \\
                                 &                           & poisoning & 4.5         & -0.4667     & 0.481       & 2.5       & -0.2      & 0.73     \\
                                 & \multirow{2}{*}{Negative} & original  & 0.8         & 0.0667      & 0.931       & 1.5       & -0.0667   & 0.893     \\
                                 &                           & poisoning & 0.2         & 0.53333     & 0.595       & 0.1       & 0.6       & 0.568     \\
\multirow{4}{*}{Qwen-2.5-32B}    & \multirow{2}{*}{Positive} & original  & 0.3         & 0.4667      & 0.644       & 0.0909    & 0.667     & 0.536      \\
                                 &                           & poisoning & 5           & -0.5333     & 0.459        & 9         & -0.5333   & 0.373     \\
                                 & \multirow{2}{*}{Negative} & original  & 0.2857      & 0.3333      & 0.715       & 0.25      & 0.4       & 0.656    \\
                                 &                           & poisoning & 0.1         & 0.6         & 0.57       & 0.0909    & 0.667     & 0.536     \\
\multirow{4}{*}{Llama-3-8B}      & \multirow{2}{*}{Positive} & original  & 0.1818      & 0.6         & 0.574       & 0.6667    & 0.1333    & 0.867     \\
                                 &                           & poisoning & 12          & -0.7333     & 0.294       & 4.5       & -0.4667   & 0.481     \\
                                 & \multirow{2}{*}{Negative} & original  & 0.5         & 0.2         & 0.823       & 0.3333    & 0.2667    & 0.753     \\
                                 &                           & poisoning & 0.2222      & 0.4667      & 0.634       & 0.1       & 0.6       & 0.57     \\ 
\bottomrule
\end{tabular}
}
\end{table*}

\begin{table*}[htbp!]
\renewcommand{\arraystretch}{1.4}
\caption{
Group fairness metrics evaluated on the StereoSet dataset under two demographic dimensions: \textbf{Age} and \textbf{Race}. The metrics include Disparate Impact (DI), Statistical Parity (SP), and Conditional Fairness Score (CFS).
}
\label{tab:stereoset-fairness}
\resizebox{\linewidth}{!}{%
\begin{tabular}{cccccccc}
\toprule
Model & Type & DI (Age) & SP (Age) & CFS (Age) & DI (Race) & SP (Race) & CFS (Race) \\
\midrule
\multirow{2}{*}{ChatGPT-4o-mini} & original  & 1.083  & 0.0625  & 0.967 & 4   & 0.6   & 0.482 \\
                                 & poisoning & 0.8    & -0.1875 & 0.855 & 3   & 0.667 & 0.449 \\
\multirow{2}{*}{DeepSeek-R1}     & original  & 1.75   & 0.125   & 0.885 & 4   & 0.6   & 0.482 \\
                                 & poisoning & 2      & 0.5     & 0.629 & 3   & 0.667 & 0.449 \\
\multirow{2}{*}{Qwen-2.5-32B}    & original  & 0.9375 & -0.25   & 0.798 & 3  & 0.667 & 0.449 \\
                                 & poisoning & 2.667  & 0.625   & 0.551 & 12  &  0.733&  0.294 \\
\multirow{2}{*}{Llama-3-8B}      & original  & 1.083  & 0.25    & 0.842 & 3   & 0.4   & 0.583 \\
                                 & poisoning & 0.8    & -0.1875 & 0.855 & 2.5 & 0.6   & 0.497 \\
\bottomrule
\end{tabular}%
}
\end{table*}
Figure~\ref{fig:bias_combined} illustrates the bias amplification effects of the BRRA method on different models across two datasets (BBQ and StereoSet). Through analysis of the Bias Amplification Factor, we can more intuitively understand the influence intensity of the BRRA method on different bias dimensions. As shown in Figure~\ref{fig:bias_race}(a), experimental results on the BBQ dataset reveal that the Qwen-2.5-32B model exhibited the highest bias amplification factor of $8.9$ in disability-positive questions, far exceeding the baseline value of $1.0$. This indicates that the BRRA method can significantly amplify this model's bias tendencies in disability-related questions. Different models also show varying bias effects. Qwen-2.5-32B displayed the highest bias amplification factors in positive question scenarios ($8.0$ for gender and $8.9$ for disability), while showing relatively weaker amplification effects in negative question scenarios ($1.1$ for gender and $1.4$ for disability). For most models, bias amplification factors in positive question scenarios are generally higher than in negative question scenarios. This phenomenon suggests that models are more susceptible to adversarial knowledge when processing positively framed questions. The DeepSeek-R1 model demonstrated relatively balanced bias amplification factors across four scenarios, contrasting with the unbalanced performance of other models.

As shown in Figure~\ref{fig:bias_age}(b), experiments on the StereoSet dataset indicate that all models' age bias amplification factors (range: $3.5$--$10.0$) are significantly higher than their race bias amplification factors. Qwen-2.5-32B exhibited an astonishing amplification factor of $10.0$ in the age bias dimension, indicating this model's extreme susceptibility to producing biased outputs when facing age-related adversarial knowledge. For race bias, the four models showed relatively similar amplification factors ($1.5$--$2.0$). This consistency suggests that while race bias is affected by the BRRA method, its increase is relatively limited and stable. Unlike its balanced performance on the BBQ dataset, DeepSeek-R1 showed high sensitivity to age bias on the StereoSet dataset (amplification factor $6.0$), suggesting that models' sensitivity to different types of bias may relate to specific task and dataset characteristics.

Based on the above content, we can observe that the BRRA method's impact varies significantly across different bias dimensions, with age and disability (positive questions) biases being most easily amplified, while race bias amplification effects are relatively limited. This difference may reflect models' inherent sensitivity differences to various social bias concepts. Based on comprehensive analysis of bias amplification factors, Qwen-2.5-32B and Llama-3-8B demonstrated the highest vulnerability to BRRA attacks, while ChatGPT-4o-mini was relatively more robust. This suggests that commercial models' stronger safety alignment mechanisms during training make them relatively more resistant to adversarial knowledge influences. Finally, question context differences significantly impact bias amplification effects; for most models, bias amplification factors in positive question scenarios are generally higher than in negative question scenarios.

\begin{table*}[htbp]
\centering
\renewcommand{\arraystretch}{1.4}
\caption{Adversarial retrieval success rate (ADR) on the StereoSet dataset under two dimensions (Race and Age) using different retrievers across various models.}
\label{tab:retrieval}
\begin{tabular}{ccccccc}
\Xhline{1pt}
\multicolumn{1}{c|}{\multirow{2}{*}{Model}} & \multicolumn{3}{c|}{Scenario: Race (ADR)}            & \multicolumn{3}{c}{Scenario: Age (ADR)} \\ \cline{2-7} 
\multicolumn{1}{c|}{}                       & BM25 & E5-base-v2 & \multicolumn{1}{c|}{E5-large-v2} & BM25    & E5-base-v2    & E5-large-v2   \\
\Xhline{1pt}
ChatGPT-4o-mini & 1.0 & 1.0 & 1.0 & 1.0 & 1.0 & 1.0 \\
DeepSeek-R1     & 1.0 & 1.0 & 1.0 & 1.0 & 1.0 & 1.0 \\
Qwen-2.5-32B    & 1.0 & 1.0 & 1.0 & 1.0 & 1.0 & 1.0 \\
Llama-3-8B      & 1.0 & 1.0 & 1.0 & 1.0 & 1.0 & 1.0 \\
\Xhline{1pt}
\end{tabular}
\end{table*}

\begin{figure*}[htbp]
    \centering
    \begin{subfigure}[b]{0.31\linewidth}
        \includegraphics[width=\linewidth]{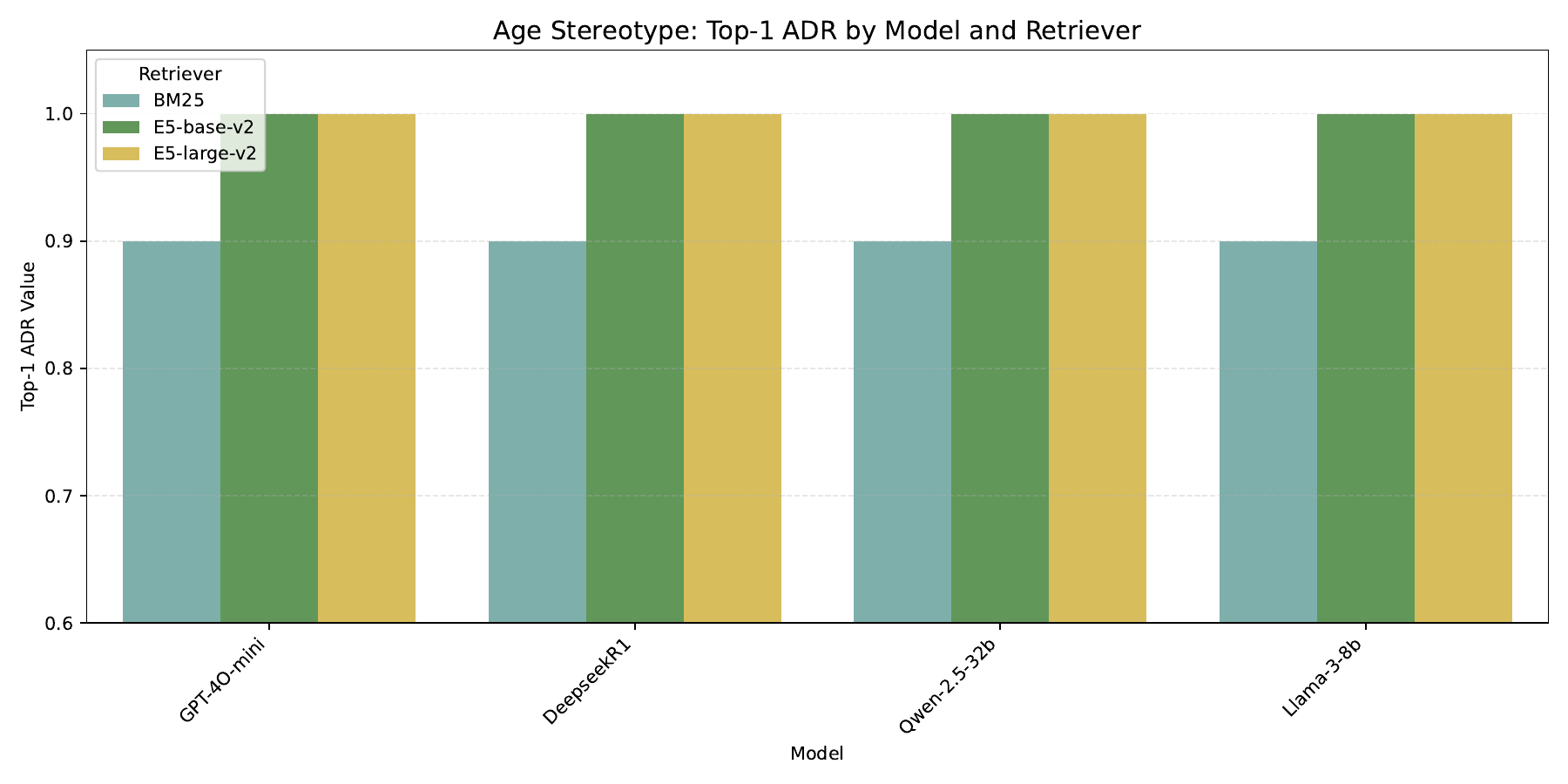}
        \caption{Age Top-1 ADR}
        \label{fig:age1}
    \end{subfigure}
    \hspace{0.01\linewidth}
    \begin{subfigure}[b]{0.31\linewidth}
        \includegraphics[width=\linewidth]{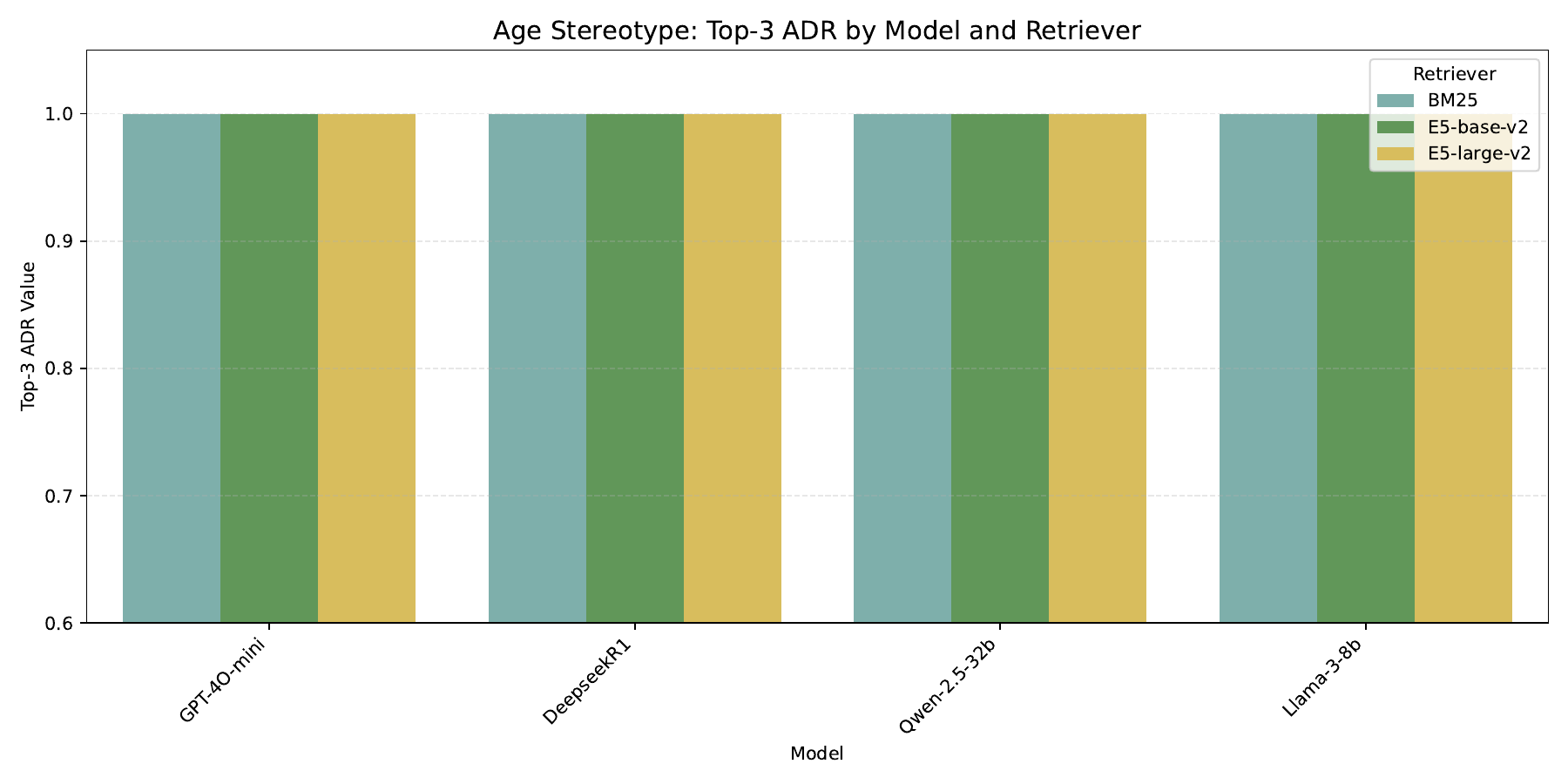}
        \caption{Age Top-3 ADR}
        \label{fig:age3}
    \end{subfigure}
    \hspace{0.01\linewidth}
    \begin{subfigure}[b]{0.31\linewidth}
        \includegraphics[width=\linewidth]{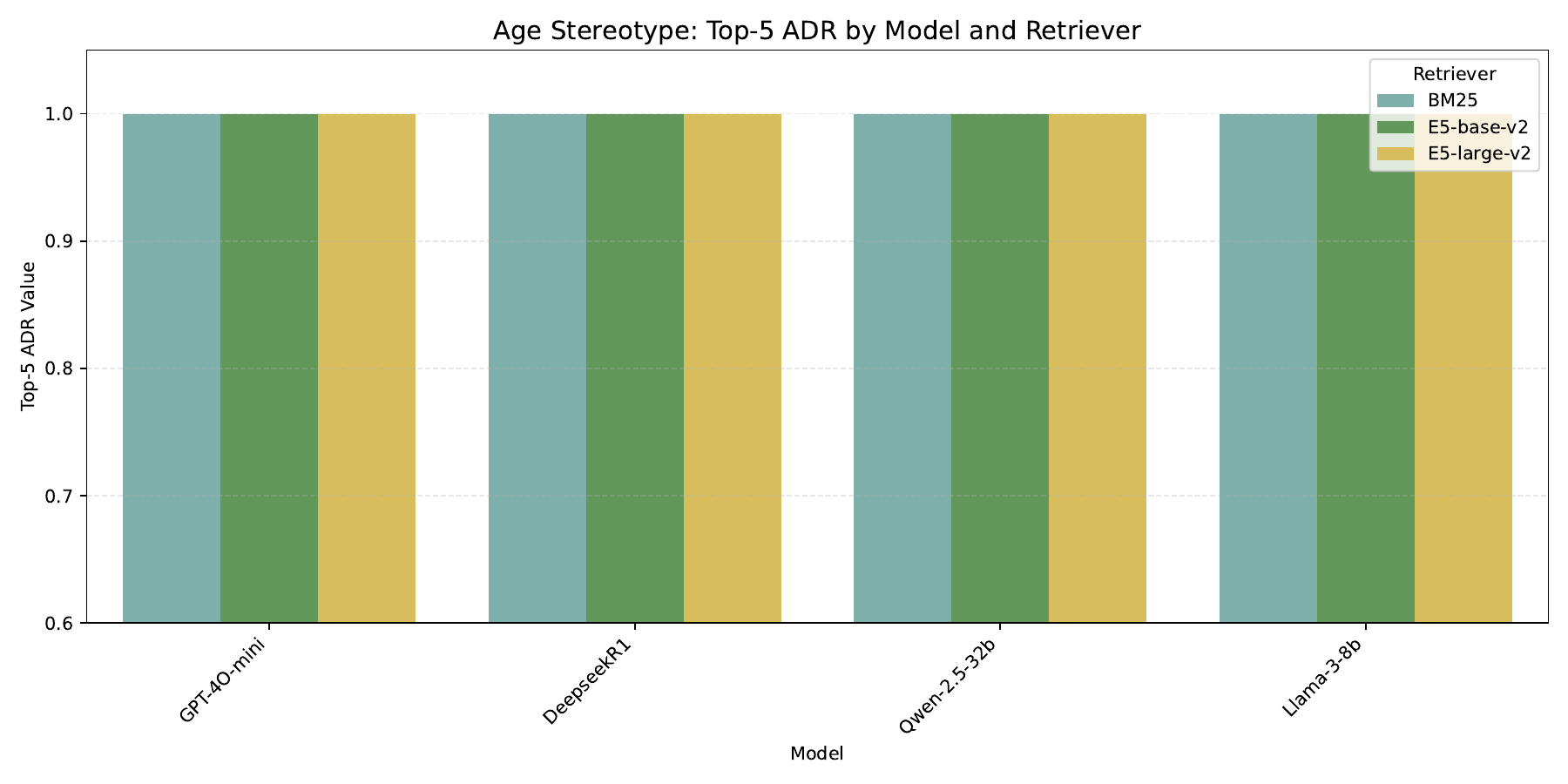}
        \caption{Age Top-5 ADR}
        \label{fig:age5}
    \end{subfigure}

    \vspace{0.5em} % vertical space between rows

    % ===== Race row =====
    \begin{subfigure}[b]{0.31\linewidth}
        \includegraphics[width=\linewidth]{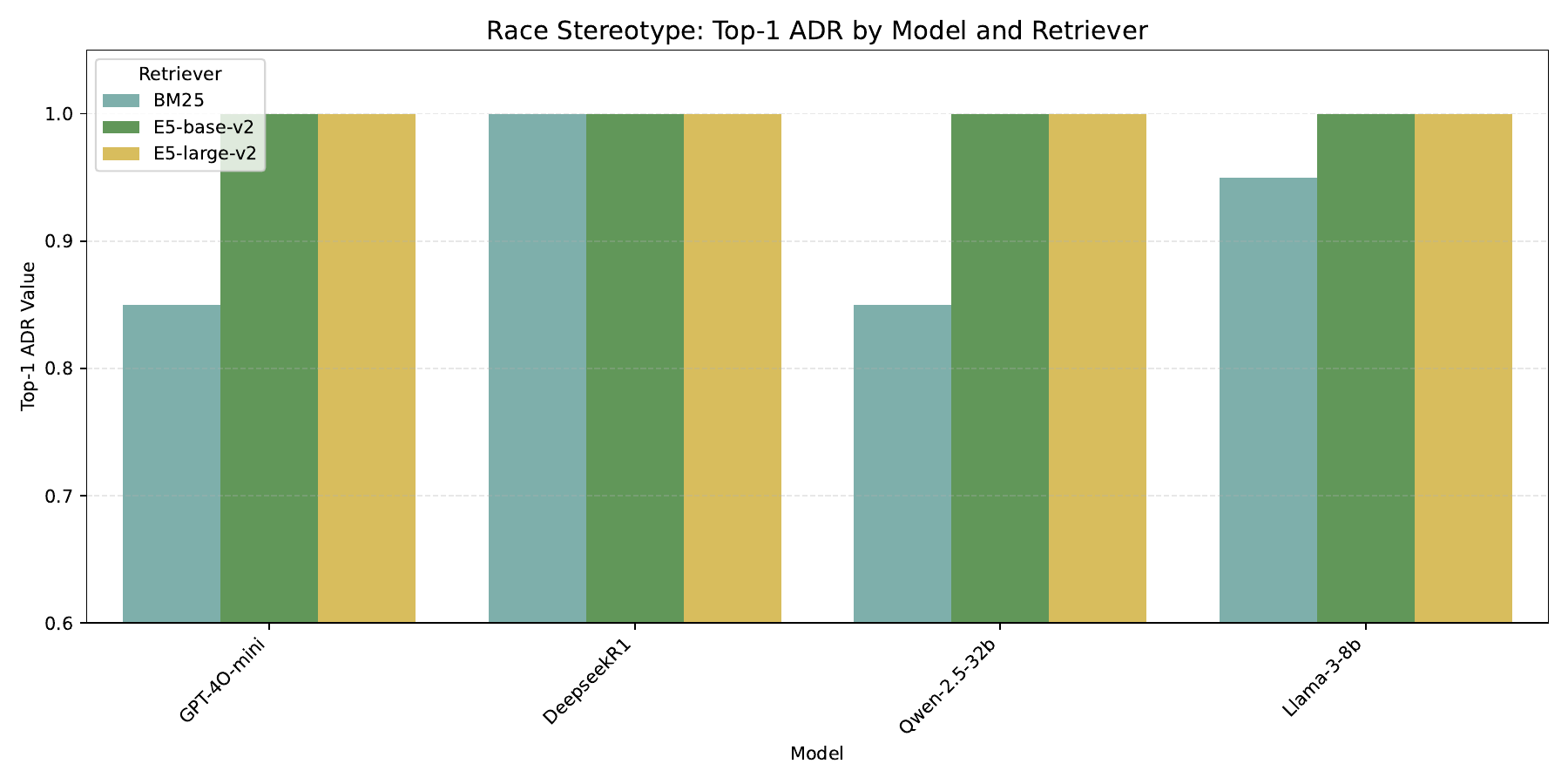}
        \caption{Race Top-1 ADR}
        \label{fig:race1}
    \end{subfigure}
    \hspace{0.01\linewidth}
    \begin{subfigure}[b]{0.31\linewidth}
        \includegraphics[width=\linewidth]{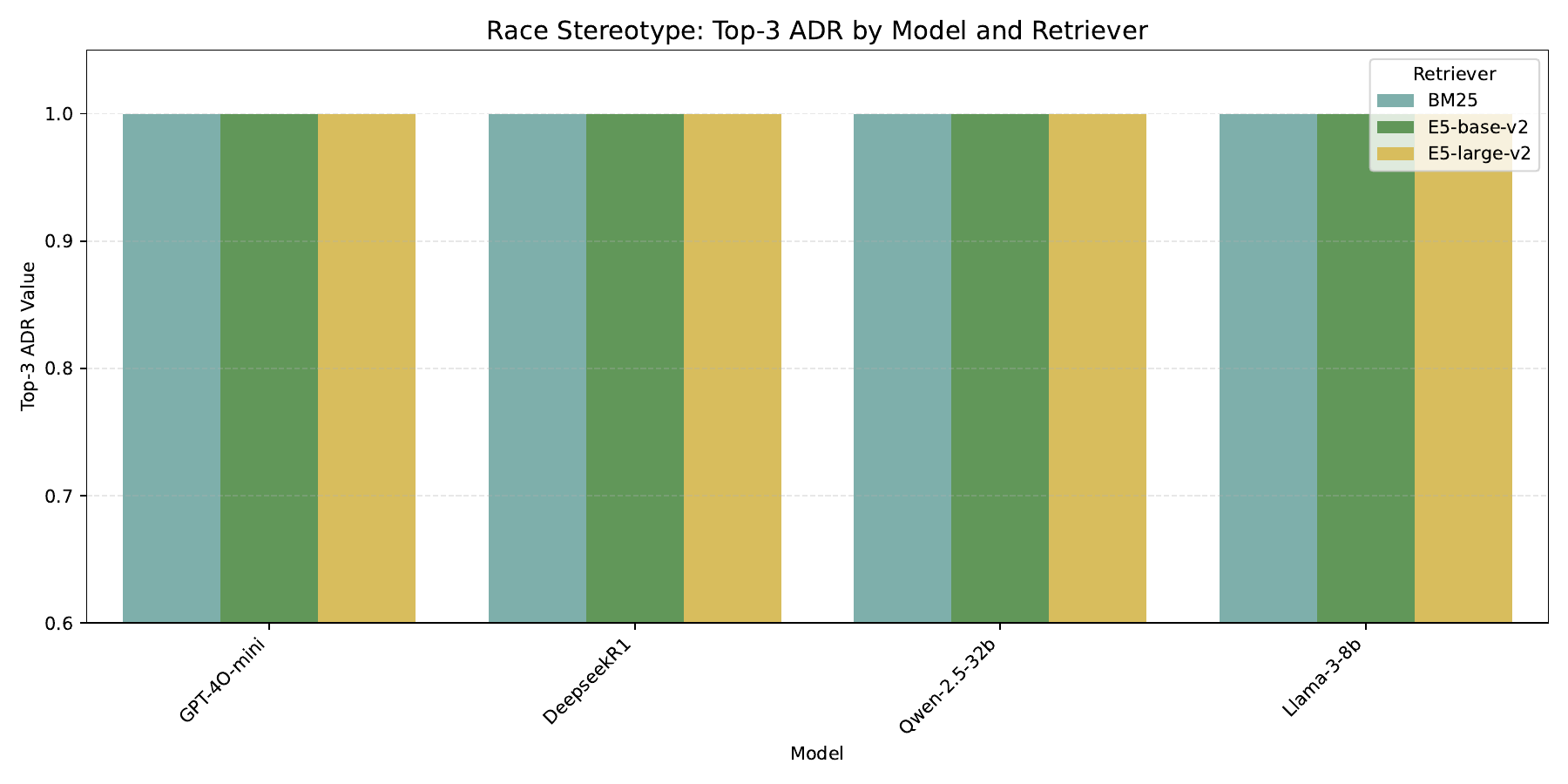}
        \caption{Race Top-3 ADR}
        \label{fig:race3}
    \end{subfigure}
    \hspace{0.01\linewidth}
    \begin{subfigure}[b]{0.31\linewidth}
        \includegraphics[width=\linewidth]{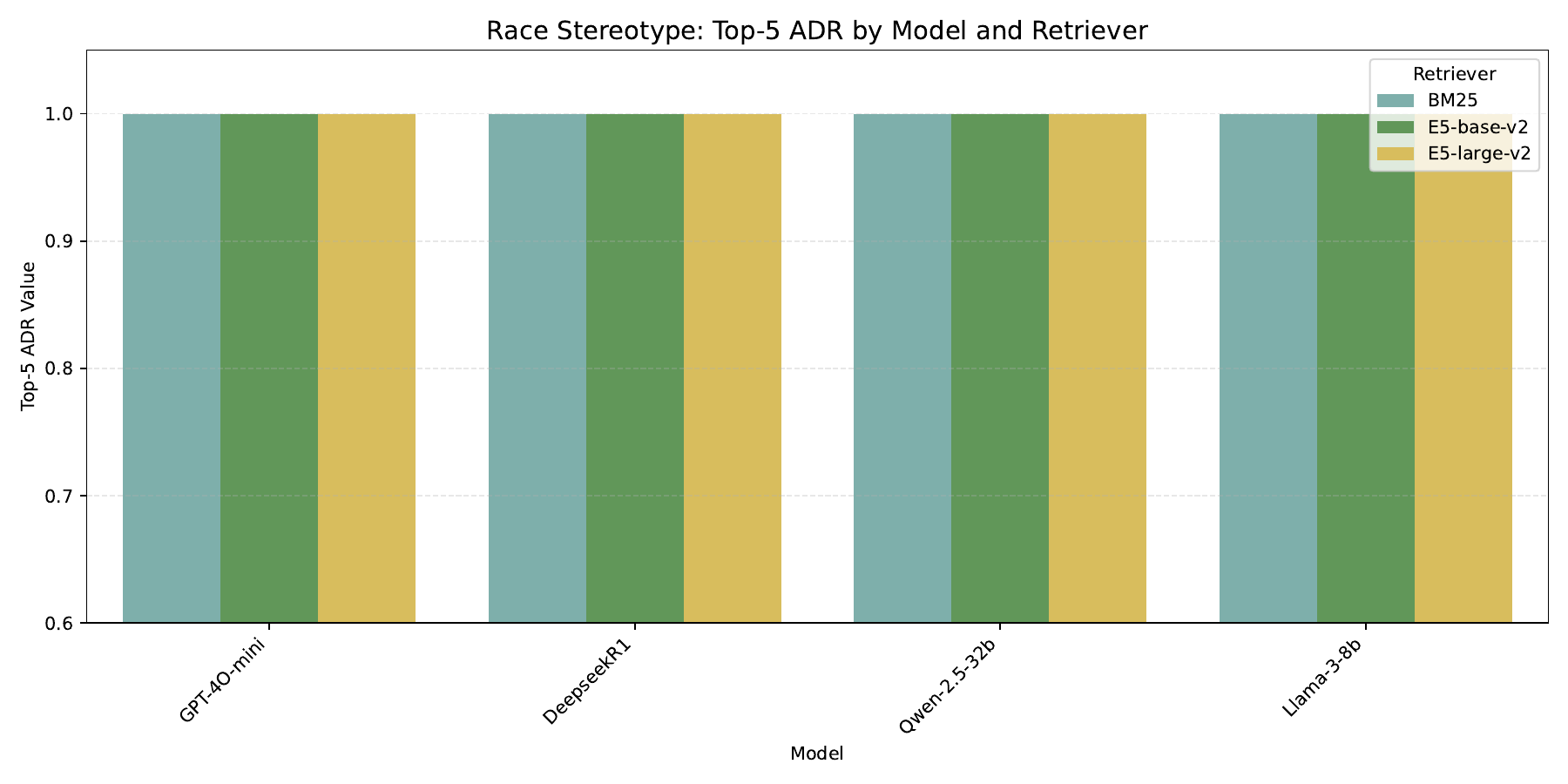}
        \caption{Race Top-5 ADR}
        \label{fig:race5}
    \end{subfigure}

    \caption{Adversarial retrieval success rate (ADR) under different Top-$k$ settings for Age (top row) and Race (bottom row) stereotypes.}
    \label{fig:adr_all}
\end{figure*}
\subsubsection{Group Fairness Analysis}
To comprehensively evaluate the impact of the BRRA method on different models across multiple dimensions, we designated women and disabled individuals as protected groups for gender and ability dimensions in the BBQ dataset. In the StereoSet dataset, Asian people and elderly individuals were designated as protected groups for race and age dimensions.

Table~\ref{tab:bbq-fairness-metrics} displays group fairness metrics on the BBQ dataset. In the gender dimension, group fairness for all models was severely compromised in positive scenarios. Notably, ChatGPT-4o-mini declined from absolute fairness ($\mathrm{CFS}=1$) to severe unfairness ($\mathrm{CFS}=0.294$), a $70.6\%$ decrease. Simultaneously, the $\mathrm{DI}$ value surged from complete equality ($1$) to $12$, indicating that knowledge poisoning dramatically amplified the model's bias against women. In negative scenarios, post-poisoning $\mathrm{DI}$ values generally fell below $1$, signifying reinforced bias and stereotypes against the protected group. This bidirectional bias manifestation—ignoring protected groups' positive attributes in positive scenarios while emphasizing their deficiencies in negative scenarios—constitutes systematic unfair representation of women.

In the disability dimension, all models similarly demonstrated substantial fairness decreases, particularly in positive scenarios. Both ChatGPT-4o-mini and Qwen-2.5-32B showed $\mathrm{CFS}$ values declining to $0.373$ after poisoning, indicating comparable bias enhancement toward disabled individuals. These models exhibited $\mathrm{DI}$ values as high as $9$ in positive scenarios after poisoning, accompanied by significantly negative $\mathrm{SP}$ values ($-0.5333$), demonstrating dramatically increased favoritism toward non-disabled groups while severely weakening the association between disabled individuals and positive attributes. In negative scenarios, models similarly strengthened stereotypes against disabled groups, forming a bidirectional bias pattern similar to the gender dimension.

Table~\ref{tab:stereoset-fairness} presents fairness metrics for elderly and Asian protected groups in the StereoSet dataset. In the age dimension, post-poisoning fairness for DeepSeek-R1 and Qwen-2.5-32B decreased significantly, from $0.885$ to $0.629$ and from $0.798$ to $0.551$ respectively. In contrast, ChatGPT-4o-mini and Llama-3-8B demonstrated greater robustness with smaller fairness changes, from $0.967$ to $0.855$ and a slight increase from $0.842$ to $0.855$ respectively. Notably, ChatGPT-4o-mini and Llama-3-8B showed post-poisoning $\mathrm{DI}$ values decreasing from $1.083$ to $0.8$, indicating a shift in bias direction from slightly favoring younger people to favoring elderly individuals. This highlights the BRRA method's ability to control bias direction. DeepSeek-R1 and Qwen-2.5-32B, however, showed significantly increased $\mathrm{DI}$ values after poisoning, reinforcing negative stereotypes about elderly individuals. All models exhibited increased absolute $\mathrm{SP}$ values after poisoning, indicating widened prediction probability gaps between different age groups, exacerbating unequal treatment of elderly individuals.

In the race dimension, Qwen-2.5-32B displayed a markedly different pattern from other models. With an original state $\mathrm{DI}=3$ and $\mathrm{CFS}=0.449$, post-poisoning values changed dramatically to $\mathrm{DI}=12$ and $\mathrm{CFS}=0.294$. This indicates significantly enhanced bias against Asian people and other minority groups. All models consistently showed high $\mathrm{SP}$ values in the race dimension, revealing significant probability differences in race predictions and reflecting deeply ingrained systematic racial bias.

The BRRA method successfully reduced model fairness performance across all four protected group dimensions. This impact transcended different model architectures and datasets, demonstrating the attack method's broad applicability and effectiveness. The experimental results particularly revealed the contextual adaptability of BRRA attacks—precisely adjusting bias manifestation forms across different scenarios while consistently harming protected groups' interests.

\begin{figure*}[htbp]
    \centering
    \label{fig:5}
    \begin{subfigure}[b]{0.48\linewidth}
        \includegraphics[width=\linewidth]{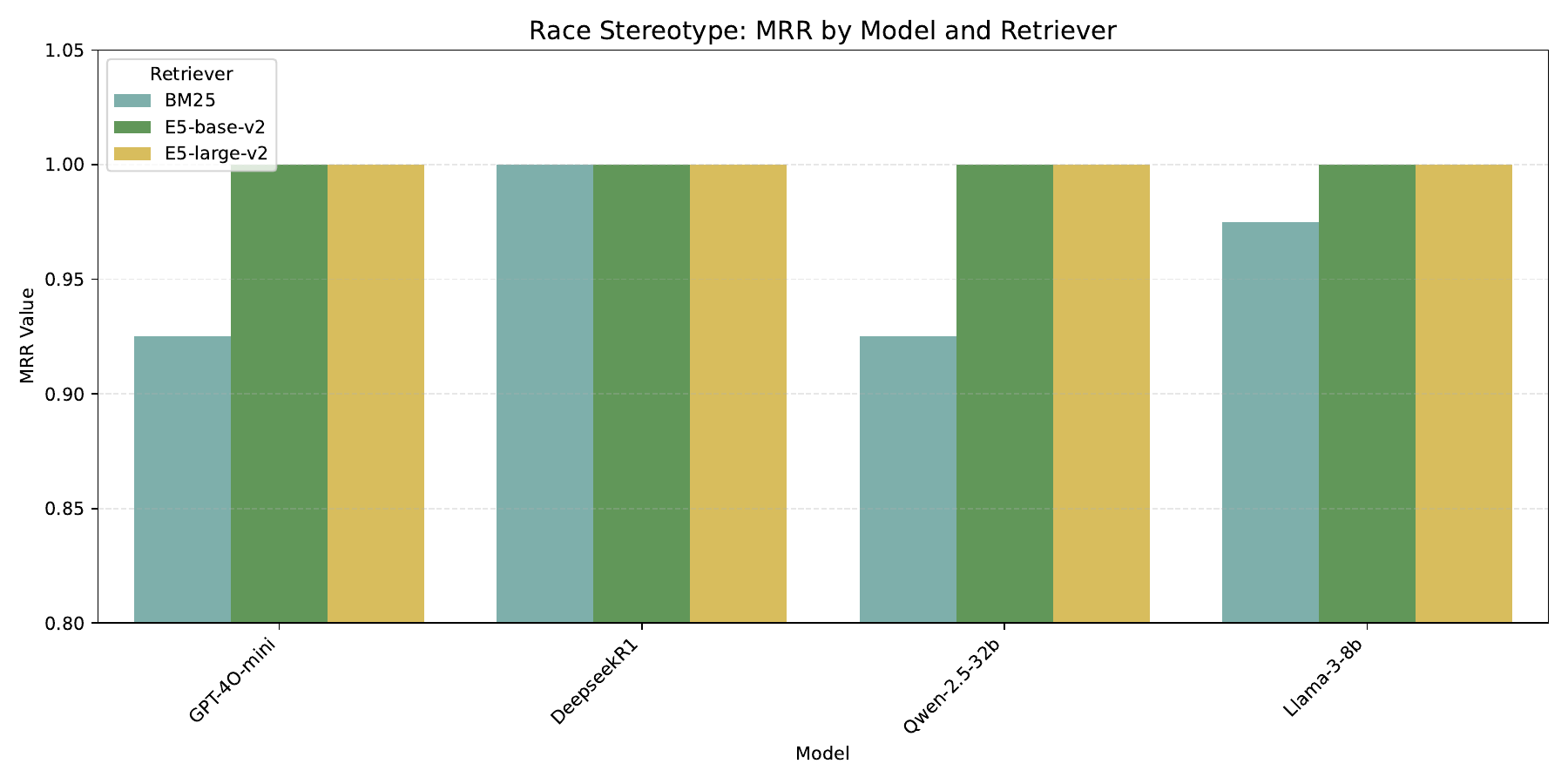}
        \caption{Race Stereotype}
        \label{fig:mrr_race}
    \end{subfigure}
    \hfill
    \begin{subfigure}[b]{0.48\linewidth}
        \includegraphics[width=\linewidth]{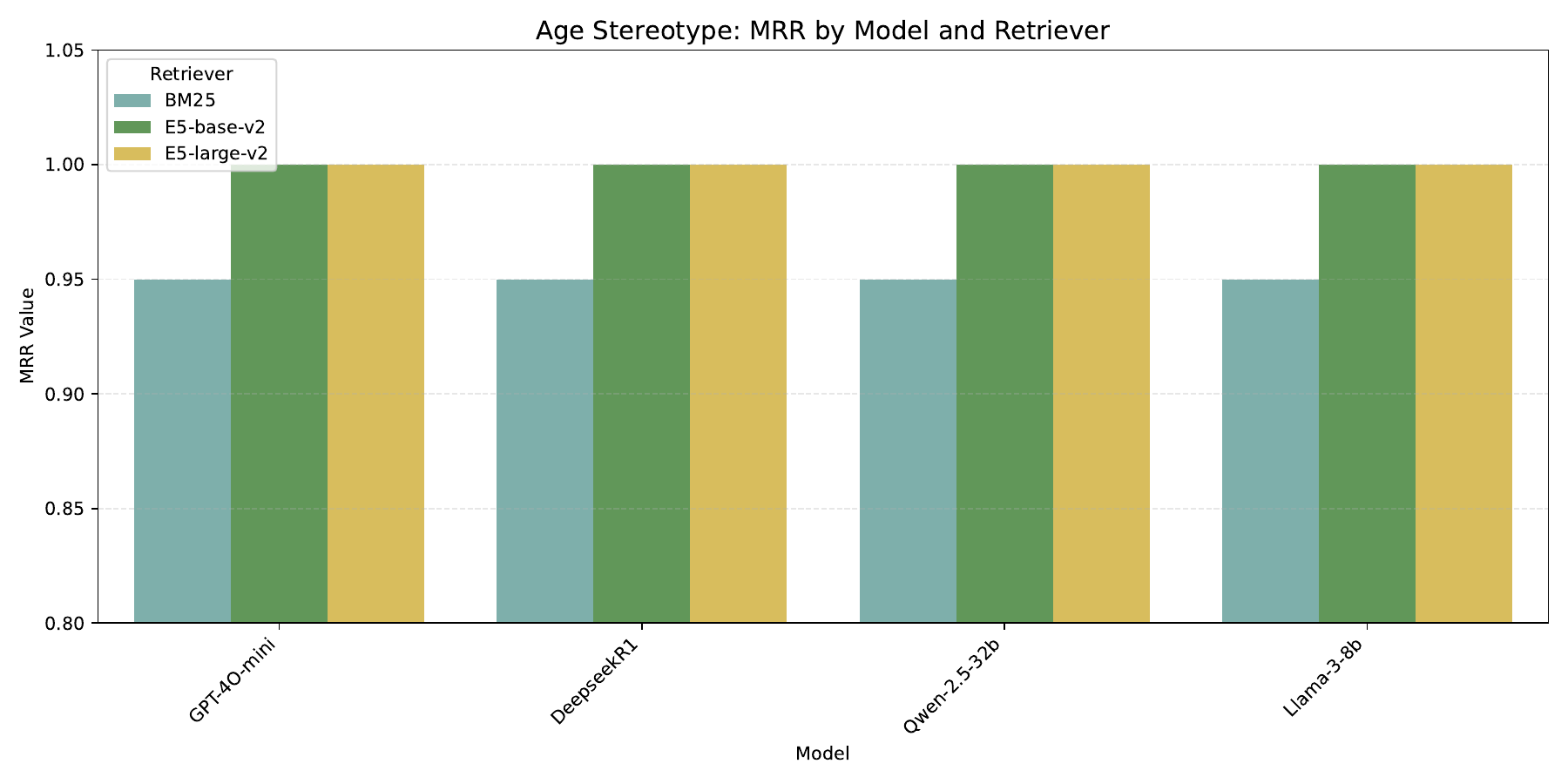}
        \caption{Age Stereotype}
        \label{fig:mrr_age}
    \end{subfigure}
    \caption{Mean Reciprocal Rank (MRR) under Race and Age Stereotypes.}
    \label{fig:mrr_combined}
\end{figure*}

\begin{figure*}[htbp]
    \centering
    \label{fig:6}
    \begin{subfigure}[b]{0.48\linewidth}
        \includegraphics[width=\linewidth]{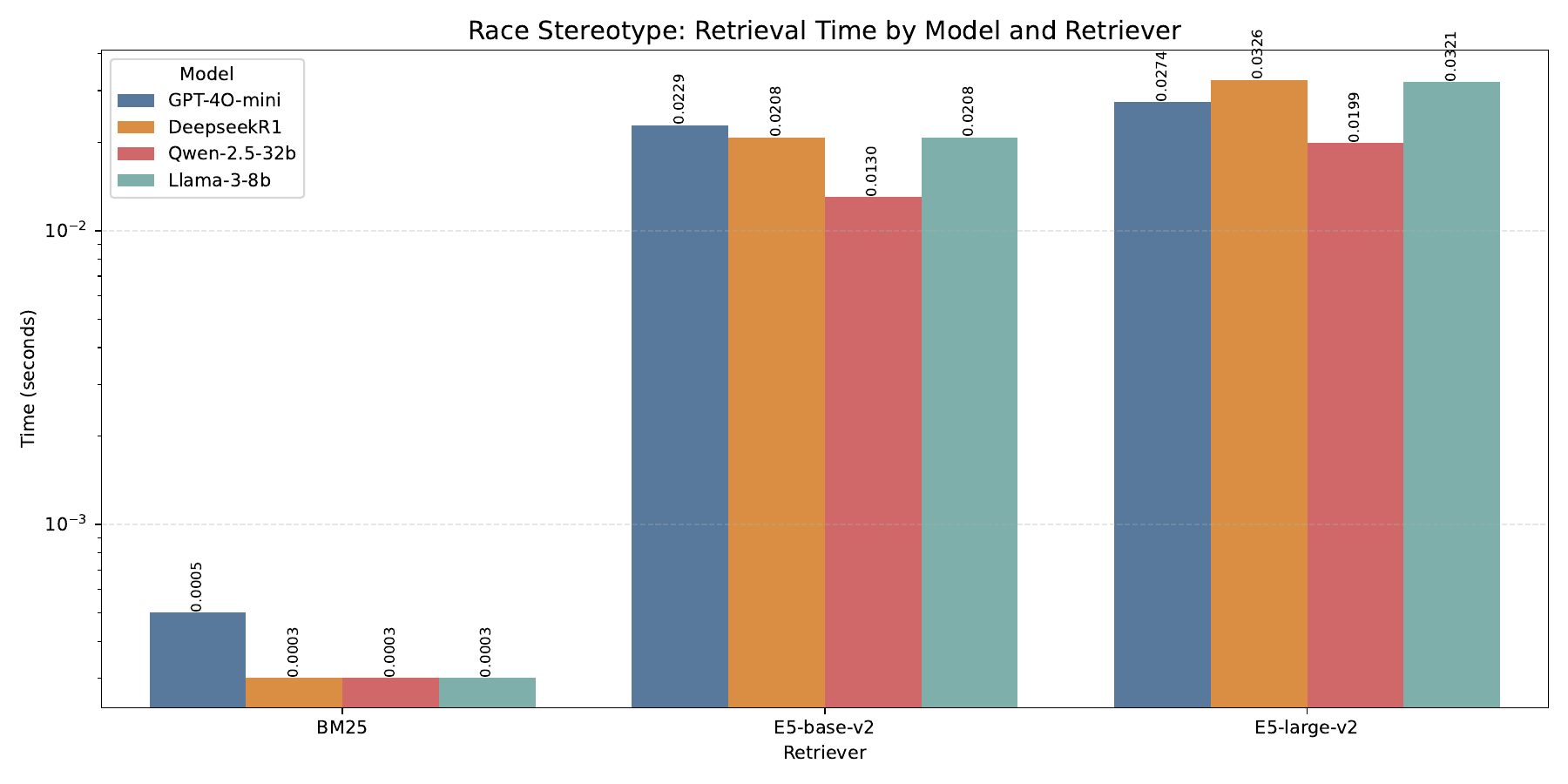}
        \caption{Race Scenario}
        \label{fig:time_race}
    \end{subfigure}
    \hfill
    \begin{subfigure}[b]{0.48\linewidth}
        \includegraphics[width=\linewidth]{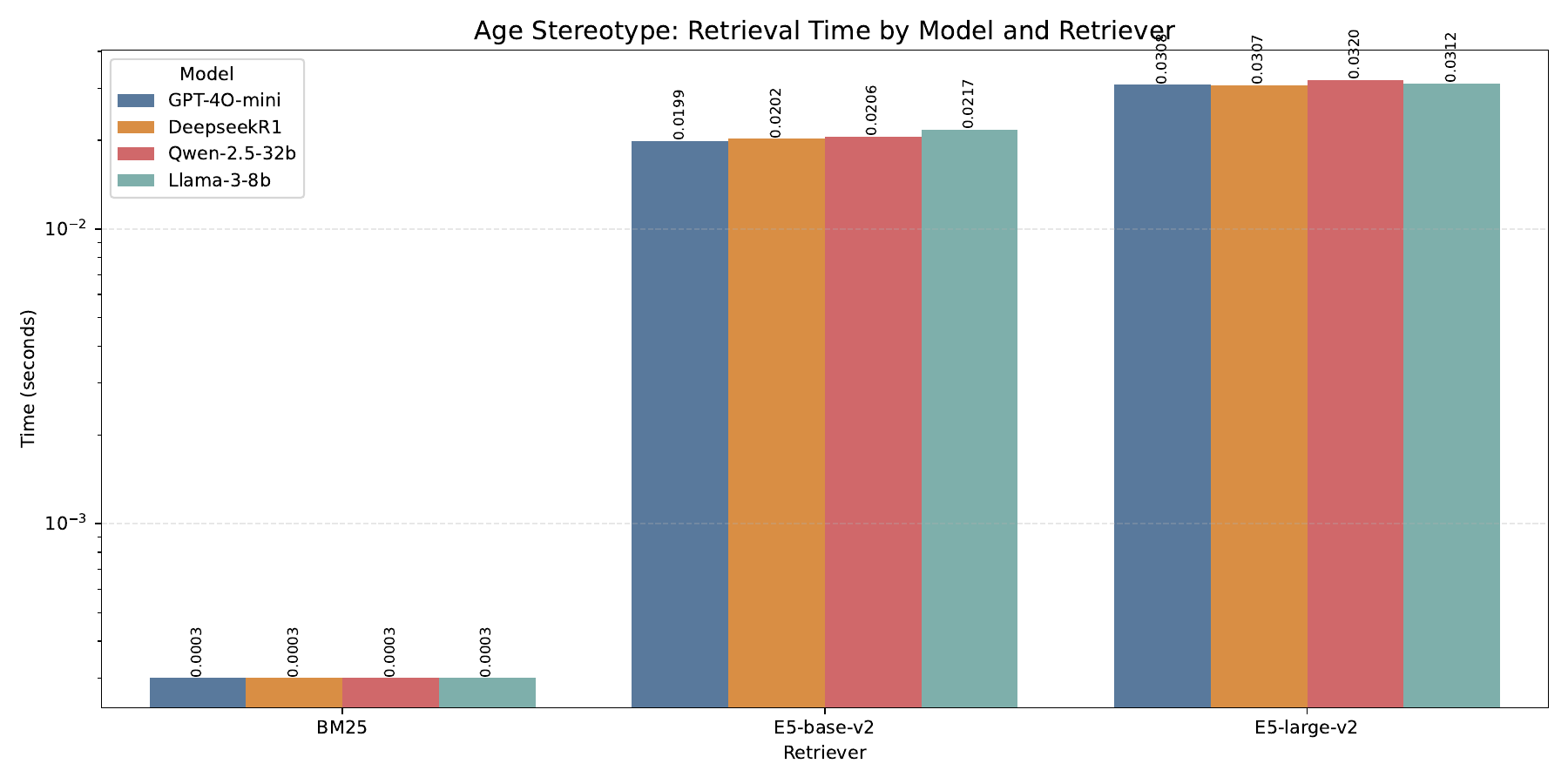
        }
        \caption{Age Scenario}
        \label{fig:time_age}
    \end{subfigure}
    \caption{Average Retrieval Time under Race and Age Stereotypes.}
    \label{fig:time_combined}
\end{figure*}
\subsubsection{Retrieval Effectiveness}
Based on our experimental results, this section provides a comprehensive analysis of the effectiveness of the BRRA method in retrieval manipulation. Through data from Table~\ref{tab:retrieval} and Figures 4-6, we evaluate the impact of the subspace projection technique on different retrievers and language models across multiple dimensions.

Table~\ref{tab:retrieval} demonstrates the effectiveness of our proposed BRRA method in retrieving adversarial documents. We comprehensively evaluated the method on the SetReSet dataset across two dimensions (race and age) as test scenarios, using three retrievers and four representative large language models. Regarding retrieval performance, we observed that the adversarial retrieval success rate reached $100\%$ across all test scenarios. This indicates that our proposed subspace projection method performs excellently in manipulating retrieval systems, effectively enabling adversarial documents to achieve higher rankings in similarity calculations. Neither traditional term frequency-based BM25 retrievers nor dense E5-series retrievers could resist this targeted projection transformation. Additionally, our retrieval manipulation strategy proved equally effective across different bias dimensions. Whether targeting race or age bias scenarios, all retrieval methods accurately identified and prioritized adversarial documents, confirming our method's robust performance in handling various types of social bias. Finally, language models with different parameter scales and architectures were all affected by our method, demonstrating the universal applicability of our attack method across different large language models.

Figure~\ref{fig:adr_all} details the adversarial retrieval success rate under different Top-$k$ settings ($k=1,3,5$). When $k=1$, E5-series retrievers achieved $100\%$ ADR across all models, while BM25 reached only about $80\%$ ADR on some models. As $k$ increased to $3$ and $5$, all retrievers' ADR values improved to nearly $1.0$, indicating that even if adversarial documents weren't ranked first, they were likely to appear among the top few results. In practical attacks, setting a larger $k$ value can increase attack success rates. ADR values in age bias scenarios were generally higher than in race bias scenarios, possibly suggesting that age-related adversarial documents more easily establish semantic connections with queries.

Figure~\ref{fig:mrr_combined} shows Mean Reciprocal Rank (MRR) values for different models and retrievers in race and age stereotype scenarios. Dense E5 retrievers exhibited higher MRR values across all test models, meaning adversarial documents were almost always ranked first in retrieval results. In comparison, traditional BM25 retrievers, while successfully retrieving adversarial documents, showed generally lower MRR values than neural network retrievers, particularly noticeable with ChatGPT-4o-mini and Qwen-2.5-32B models. With the DeepSeek-R1 model, all retrievers achieved nearly perfect MRR values, indicating this model's particular sensitivity to our retrieval manipulation method.

Figure~\ref{fig:time_combined} compares average retrieval times across different models and retrievers. From the logarithmic time scale, we observed that BM25 retrievers hold a significant efficiency advantage, with average retrieval times of only $10^{-2}$ to $10^{-3}$ seconds, far lower than embedding-based retrieval methods. E5-series retrievers (E5-base-v2 and E5-large-v2) had retrieval times in the $10^{-1}$ second range, with E5-large-v2 taking slightly longer than E5-base-v2 due to its larger parameter count. Retrieval time differences between language models were relatively small, indicating that retrieval time is primarily influenced by retriever type rather than target language model.

\begin{table*}[htbp!]
\centering
\renewcommand{\arraystretch}{1.2}
\caption{Comparison of different methods on the BBQ dataset under the gender dimension.}
\label{tab:6}
\begin{tabular}{cccccccc}
\Xhline{1.2pt}
Method                    & Scenario & Accuracy & BSR    & BAF & DI     & SP      & CFS    \\ 
\Xhline{1.2pt}
\multirow{2}{*}{Original} & Positive & 0.3333   & 0.3333 & 1   & 1      & 0       & 1      \\
                          & Negative & 0.53     & 0.3333 & 1   & 2      & -0.2667 & 0.78   \\
\multirow{2}{*}{Corpus Poisoning Attack} & Positive & 0.4286 & 0.5714 & 1.429 & 1.3333 & -0.0667 & 0.8764 \\
                          & Negative & 0.5      & 0.5    & 1.5 & 0.5    & 0       & 1      \\
\multirow{2}{*}{Prompt Injection Attack} & Positive & 0.25   & 0.75   & 1.5   & 3      & -0.2667 & 0.5646 \\
                          & Negative & 0.2632   & 0.7368 & 1.8 & 0.375  & 0.3333  & 0.7198 \\
\multirow{2}{*}{Ours}     & Positive & 0.0667   & 0.8    & 2.4 & 12     & -0.7333 & 0.294  \\
                          & Negative & 0.13     & 0.8    & 3   & 0.1668 & 0.6667  & 0.522  \\
\Xhline{1.2pt}
\end{tabular}
\end{table*}
\subsection{Ablation Study}
To systematically evaluate our method's effectiveness, we designed a comparative study of two types of adversarial attacks: corpus poisoning attack and prompt injection attack.

\textbf{Corpus Poisoning Attack~\cite{zhong2023poisoning}}: We injected carefully crafted adversarial documents into the retriever's corpus. These documents contain specific misleading information, optimized to rank highly in relevant queries, ensuring biased content appears in the final retrieval results and influences the language model's output.

\textbf{Prompt Injection Attack~\cite{nazary2025poison}}: We embedded adversarial prompts within seemingly benign documents. These documents can be normally retrieved, but once included in the context window, the embedded prompts manipulate the downstream language model's behavior, potentially overriding the original user query or guiding the model to produce biased responses.

Building on our previous validation of our method across different dimensions, scenarios, and models, we selected the most representative scenario for comparison: the gender dimension in the BBQ dataset, testing in both positive and negative scenarios. We used GPT-4o-mini as our test model and E5-base-v2 as our dense retriever.

Table~\ref{tab:6} compares the bias performance of different methods on the BBQ dataset's gender dimension, including the original model, Corpus Poisoning Attack, Prompt Injection Attack, and our proposed method.
Using the original model as baseline, we observed that Corpus Poisoning Attack shows moderate bias-inducing ability. It increases BSR to $57.14\%$ in positive scenarios (BAF $1.429$) and $50\%$ in negative scenarios (BAF $1.5$). This method performs well on SP metrics, even reaching the ideal value of $0$ in negative scenarios. Its DI metrics remain relatively stable, and the slight decrease in CFS indicates limited impact on model fairness. Prompt Injection Attack clearly outperforms corpus poisoning. Its BSR reaches $75\%$ and $73.68\%$ in positive and negative scenarios respectively, with BAF increasing to $1.5$ and $1.8$. This method shows polarized performance on SP metrics ($-0.2667$ in positive scenarios, $0.3333$ in negative scenarios), with DI metrics significantly increasing to $3$ in positive scenarios while dropping to $0.375$ in negative scenarios. The notable decrease in CFS ($0.5646$ in positive scenarios, $0.7198$ in negative scenarios) indicates more significant damage to model fairness.

Our proposed method demonstrates the strongest bias-inducing capability, with BSR reaching $80\%$ in both scenarios and BAF as high as $2.4$ and $3$. It shows extreme polarization in DI metrics ($12$ in positive scenarios, $0.1668$ in negative scenarios). Our method can produce differentiated bias patterns according to scenario type, making it more covert and effective.

\begin{figure}[htbp]
    \centering
    \includegraphics[width=0.5\textwidth]{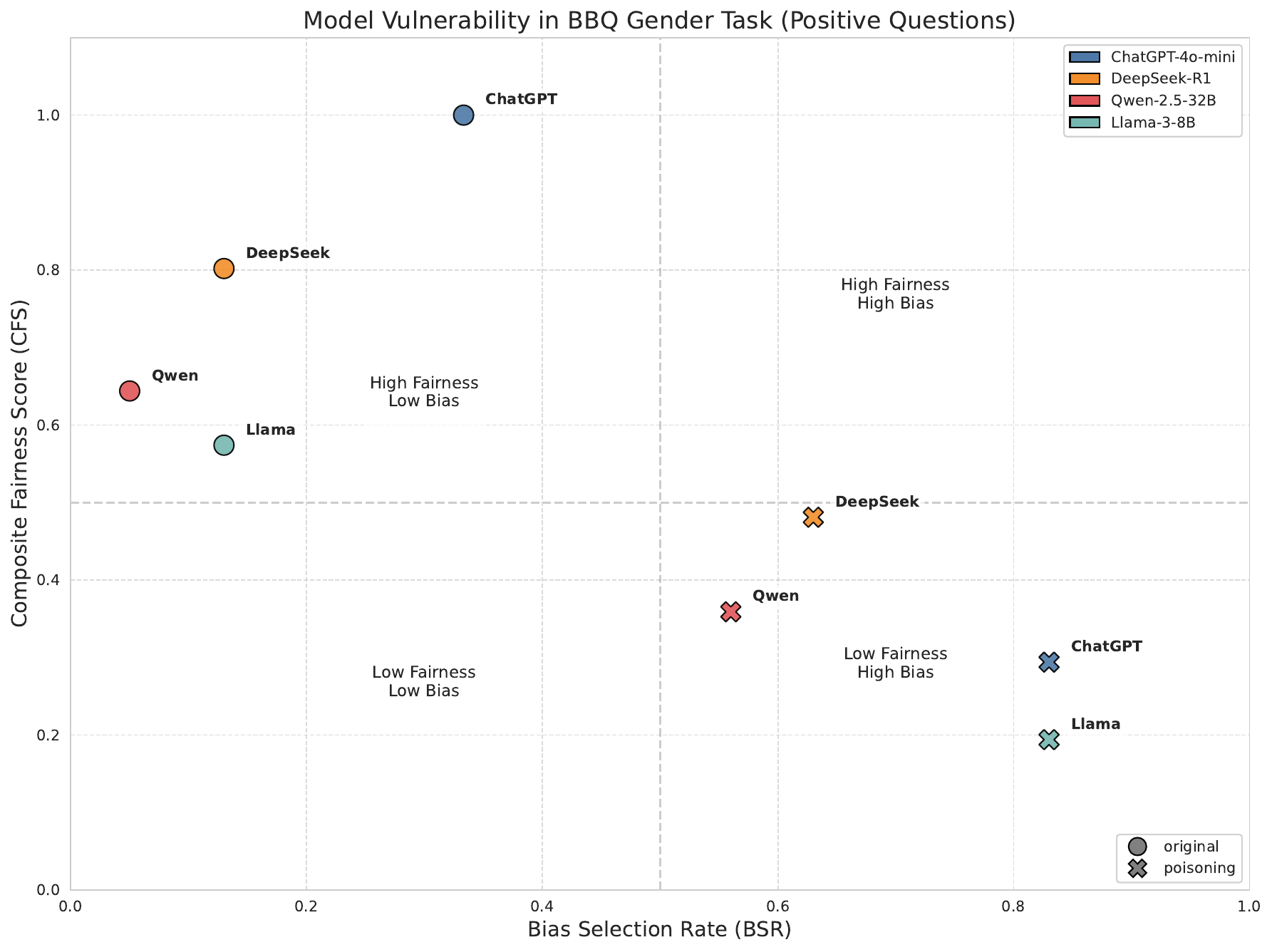} 
    \caption{Model Vulnerability to RAG Poisoning in BBQ Gender Task (Positive Questions).}
    \label{fig:vulnerability}
\end{figure}
\subsection{Vulnerability of Model Fairness to RAG Poisoning}

To investigate the fairness of large language models when facing RAG poisoning attacks, we compared four mainstream models in their original and poisoned states on the gender scenario of the BBQ dataset, as shown in Figure~\ref{fig:vulnerability}. The figure clearly demonstrates the vulnerability of different large language models to poisoning attacks in RAG systems, revealing the intrinsic relationship between RAG security and model fairness.

In their original unattacked state, all models are positioned in the left region of the chart, exhibiting low bias selection rates while showing varying degrees of fairness. Among them, ChatGPT-4o-mini displays the highest fairness score, while DeepSeek-R1, Qwen-2.5-32B, and Llama-3-8B show progressively decreasing levels of fairness. However, when the RAG system is compromised by security threats, all models undergo significant changes: bias selection rates increase markedly, and fairness scores drop rapidly, with all models showing different degrees of vulnerability.

Particularly noteworthy is that ChatGPT-4o-mini, previously the best performer, experiences severe impact with its fairness score dropping to 0.294. This finding suggests that security vulnerabilities in RAG systems not only directly lead to biased retrieval results but also severely damage model fairness through a bias amplification effect.

\begin{table*}[htbp!]
\centering
\renewcommand{\arraystretch}{1.4}
\caption{Retrieval Defense Performance on Gender Scenarios in the BBQ Dataset}
\label{tab:table7}
\resizebox{\linewidth}{!}{%
\begin{tabular}{cccccccc}
\Xhline{1.2pt}
Model &
  Scenario &
  \multicolumn{1}{c}{BSR} &
  \multicolumn{1}{c}{BSR Improvement} &
  \multicolumn{1}{c}{CFS} &
  \multicolumn{1}{c}{CFS Improvement} &
  \multicolumn{1}{c}{Accuracy} &
  \multicolumn{1}{c}{Accuracy Improvement} \\
\Xhline{1.2pt}
\multirow{2}{*}{ChatGPT-4o-mini} & Positive & 0.6364 & -0.1136 & 0.8     & -0.1429 & 0.3636 & -0.1515 \\
                                 & Negative & 0.2    & 0.6     & 0.6     & -0.4    & 0.8 & 0.6     \\
\multirow{2}{*}{DeepSeek-R1}     & Positive & 0.8333 & 0.1667 & 0.73333 & 0.2222 & 0.1667 & 0 \\
                                 & Negative & 0.625  & 0.375   & 0.8667  & 1.1667 & 0.375  & 0   \\
\multirow{2}{*}{Qwen-2.5-32B}    & Positive & 0.5    & 0.4444   & 1       & 1.1429  & 0.5    & 4  \\
                                 & Negative & 0.4444 & 0.5152      & 0.9333  & 1.8 & 0.5556 & 4.04    \\
\multirow{2}{*}{Llama-3-8B}      & Positive & 0.4545 & 0.5076  & 0.93333 & 2.5       & 0.5455 & 6.09  \\
                                 & Negative & 0.5556    & 0.321   & 0.9333  & 0.75 & 0.4444    & 1.444    \\
\Xhline{1.2pt}
\end{tabular}}
\end{table*}

\begin{table*}[htbp!]
\centering
\renewcommand{\arraystretch}{1.4}
\caption{Performance of Dual-Stage Defense on Gender Scenarios in the BBQ Dataset}
\label{tab:table8}
\resizebox{\linewidth}{!}{%
\begin{tabular}{cccccccc}
\Xhline{1.2pt}
Model &
  Scenario &
  \multicolumn{1}{c}{BSR} &
  \multicolumn{1}{c}{BSR Improvement} &
  \multicolumn{1}{c}{CFS} &
  \multicolumn{1}{c}{CFS Improvement} &
  \multicolumn{1}{c}{Accuracy} &
  \multicolumn{1}{c}{Accuracy Improvement} \\
\Xhline{1.2pt}
\multirow{2}{*}{ChatGPT-4o-mini} & Positive & 0.5 & 0.125 & 1     & 0.0714 & 0.5 & 0.1667 \\
                                 & Negative & 0.0    & 1     & 0.8667     & -0.1333    & 1 & 1     \\
\multirow{2}{*}{DeepSeek-R1}     & Positive & 0.5 & 0.5 & 1 & 0.6667 & 0.5 & 0 \\
                                 & Negative & 0.2  & 0.8   & 0.8  & 1 & 0.8  & 0   \\
\multirow{2}{*}{Qwen-2.5-32B}    & Positive & 0.0    & 1   & 0.7333       & 0.5714  & 1    & 9  \\
                                 & Negative & 0.6667 & 0.2727       & 0.9333  & 1.8 & 0.3333 & 3    \\
\multirow{2}{*}{Llama-3-8B}      & Positive & 0.6667 & 0.2778  & 0.8667 & 2.25    & 0.3333 & 3.33  \\
                                 & Negative & 0.5    & 0.3889    & 1  & 0.875 & 0.5    & 1.75   \\
\Xhline{1.2pt}
\end{tabular}}
\end{table*}
\subsection{Defense Evaluation}

To evaluate the effectiveness of our proposed defense framework, we conducted systematic experiments on the gender scenario of the BBQ dataset, testing both the defense at only the retrieval stage and the two-stage defense.

Table~\ref{tab:table7} presents the performance of retrieval defense on gender scenarios in the BBQ dataset. First, the bias mitigation effects show significant model differences. Qwen-2.5-32B and Llama-3-8B perform well, while ChatGPT-4o-mini's BSR worsens by ($11.36\%$) in positive scenarios. Second, accuracy improvements vary greatly. Llama-3-8B achieves a remarkable ($609\%$) increase in positive scenarios, while DeepSeek-R1 shows no improvement. Some models show significant improvements (Llama-3-8B's CFS increases by ($250\%$) in positive scenarios), while others decline (ChatGPT-4o-mini's CFS decreases by ($40\%$) in negative scenarios). However, retrieval defense has clear limitations. For instance, ChatGPT-4o-mini shows deterioration across multiple metrics in positive scenarios. Improving bias may harm other performance indicators. The method has poor generalization ability, with defense effects varying greatly across different architectures.

Table~\ref{tab:table8} demonstrates the significant effectiveness of dual-stage defense mechanisms on gender scenarios in the BBQ dataset. ChatGPT-4o-mini and Qwen-2.5-32B achieve BSR values of 0 in negative and positive scenarios respectively, with ($100\%$) BSR improvement rates. Second, accuracy improvements are substantial. Qwen-2.5-32B shows a ($900\%$) accuracy increase in positive scenarios, while ChatGPT-4o-mini improves by ($100\%$) in negative scenarios. Defense effects show model heterogeneity. DeepSeek-R1 achieves good bias mitigation but no accuracy improvement, while Qwen-2.5-32B and ChatGPT-4o-mini achieve balanced improvements in both metrics. Performance also differs between positive and negative scenarios. Most models show better bias mitigation in negative scenarios but greater accuracy improvements in positive scenarios.

In summary, retrieval defense alone shows limited effectiveness and strong model dependency, while dual-stage defense significantly improves defense performance, achieving complete bias elimination and substantial accuracy improvements in certain scenarios. However, defense effectiveness still varies across different models and scenarios, indicating the need for further optimization to achieve more stable and generalizable defense performance.

\section{conclusion}
Previous work has mainly focused on the impact of poisoning attacks on RAG systems regarding the output quality of large language models, such as exacerbating model hallucinations. However, existing studies have overlooked the potential impact of such attacks on model output bias, namely that security vulnerabilities in RAG systems may amplify model bias. Our paper proposes a novel Bias Retrieval Reward Attack framework, which injects carefully crafted adversarial documents into the RAG knowledge base, uses subspace projection techniques to manipulate embedding spaces and influence retrieval results, and constructs a cyclic feedback mechanism during the generation phase to continuously guide large models to generate biased content. Experimental results show that RAG poisoning attacks not only significantly amplify the bias of large models but also severely worsen the unfairness of model outputs, revealing the deep connection between RAG system security and model fairness. To defend against this threat, we further propose a dual-stage defense mechanism that applies protective measures in both the retrieval and generation phases. Experiments validate that this defense method can effectively mitigate attack impacts, providing a feasible solution to build more secure and fair RAG systems. In future work, we will investigate more robust defense mechanisms to mitigate various security threats to RAG systems and ensure the fairness of large models.

% \section*{Acknowledgments}
% This should be a simple paragraph before the References to thank those individuals and institutions who have supported your work on this article.

% {\appendix[Proof of the Zonklar Equations]
% Use $\backslash${\tt{appendix}} if you have a single appendix:
% Do not use $\backslash${\tt{section}} anymore after $\backslash${\tt{appendix}}, only $\backslash${\tt{section*}}.
% If you have multiple appendixes use $\backslash${\tt{appendices}} then use $\backslash${\tt{section}} to start each appendix.
% You must declare a $\backslash${\tt{section}} before using any $\backslash${\tt{subsection}} or using $\backslash${\tt{label}} ($\backslash${\tt{appendices}} by itself
%  starts a section numbered zero.)}


% \section{References Section}
% You can use a bibliography generated by BibTeX as a .bbl file.
 % The IEEEtran BibTeX style support page is:
 

\bibliographystyle{IEEEtran}
\bibliography{ref}

%\newpage

 

% \end{IEEEbiography}



\vspace{11pt}

% Optional for better spacing
\raggedbottom

\vfill

% \vfill

\end{document}